\RequirePackage{fix-cm}
\documentclass[twocolumn]{svjour3}          %
\smartqed  %
\usepackage{graphicx}

\usepackage{xcolor}
\usepackage{amsmath}
\usepackage{amssymb}
\usepackage{stmaryrd}
\usepackage{booktabs}
\usepackage{subcaption}
\usepackage{multirow}
\usepackage{cite}
\usepackage{siunitx}
\usepackage[absolute]{textpos}

\newcommand\MYhyperrefoptions{bookmarks=true,bookmarksnumbered=true,
pdfpagemode={UseOutlines},plainpages=false,pdfpagelabels=true,
colorlinks=true,
pdftitle={Reference Pose Generation for Visual Localization via Learned Features and View Synthesis},%
pdfsubject={Computer Vision},%
pdfauthor={Z. Zhang, T. Sattler, D. Scaramuzza},%
pdfkeywords={Visual Localization, Benchmark Construction, Learned Local Features}}%
\usepackage[\MYhyperrefoptions,pdftex,colorlinks]{hyperref}

\newcommand{\Sec}{Sect.~}
\newcommand{\Fig}{Fig.~}
\newcommand{\Tab}{Table~}

\newcommand{\eg}{e.g., }
\newcommand{\ie}{i.e., }

\newcommand{\PAR}[1]{\vskip4pt \noindent{\bf #1~}}

\newcommand{\R}{\mathtt{R}}
\global\long\def\T{\mathtt{{T}}}
\global\long\def\J{\mathtt{{J}}}
\global\long\def\t{\mathbf{{c}}}

\newcommand{\jrange}[1]{_{j=1}^{#1}}
\newcommand{\lrange}[1]{_{l=1}^{#1}}

\newcommand{\px}{\mathbf{u}}

\global\long\def\pt{\mathbf{{p}}}

\DeclareMathOperator*{\argmin}{arg\,min}

\newcommand{\img}{I}

\begin{document}
\definecolor{somegray}{rgb}{0.4, 0.4, 0.4}
\newcommand{\darkgrayed}[1]{\textcolor{somegray}{#1}}
\begin{textblock}{16}(0, 0.15)
\begin{center}
\Large{
\darkgrayed{This paper has been accepted for publication at the \\
International Journal of Computer Vision (IJCV), 2020.
\copyright Springer
}
}
\end{center}
\end{textblock}

\title{Reference Pose Generation for Long-term Visual Localization via Learned Features and View Synthesis%
}

\author{Zichao Zhang \and Torsten Sattler \and Davide Scaramuzza %
}

\institute{Z. Zhang and D. Scaramuzza \at
              Robotics and Perception Group \\
              University of Zurich\\
              \email{zzhang@ifi.uzh.ch, sdavide@ifi.uzh.ch}           %
           \and
           T. Sattler \at
           Czech Institute of Informatics, Robotics and Cybernetics\\
           Czech Technical University in Prague\\
           \email{torsten.sattler@cvut.cz}
}

\date{Received: date / Accepted: date}

\maketitle

\begin{abstract}
Visual Localization is one of the key enabling technologies for autonomous driving and augmented reality. High quality datasets with accurate 6 Degree-of-Freedom (DoF) reference poses are the foundation for benchmarking and improving existing methods. Traditionally, reference poses have been obtained via Structure-from-Motion (SfM). However, SfM itself relies on local features which are prone to fail when images were taken under different conditions, e.g., day/ night changes. At the same time, manually annotating feature correspondences is not scalable and potentially inaccurate. In this work, we propose a semi-automated approach to generate reference poses based on feature matching between renderings of a 3D model and real images via learned features. Given an initial pose estimate, our approach iteratively refines the pose based on feature matches against a rendering of the model from the current pose estimate. We significantly improve the nighttime reference poses of the popular Aachen Day-Night dataset, showing that state-of-the-art visual localization methods  perform better (up to $47\%$) than predicted by the original reference poses. We extend the dataset with new nighttime test images, provide uncertainty estimates for our new reference poses, and introduce a new evaluation criterion. We will make our reference poses and our framework publicly available upon publication.

\keywords{Visual localization, benchmark construction, learned local features}
\end{abstract}

\begin{figure*}[t]
    \centering
    \includegraphics[width=\linewidth]{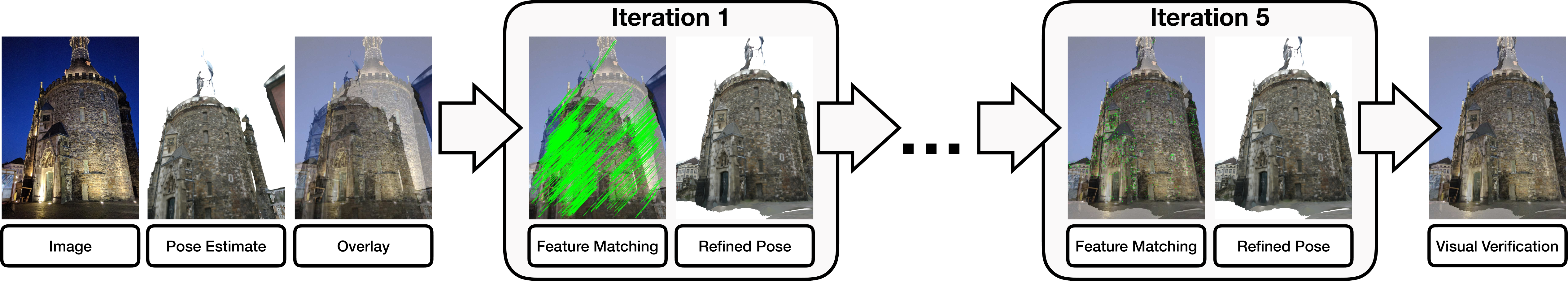}%
    \caption{Overview of our approach: Given an image, we render a synthesized view of a 3D model from the given initial pose estimate of the image. Superimposing the rendered image over the original image provides a visual cue on the accuracy of the pose estimate. We match features extracted from the actual image and the rendering (shown as green lines connecting the corresponding positions in the overlay of the two images). This provides 2D-3D correspondences between the image and the underlying scene model. These 2D-3D matches are then used to obtain a refined estimate. Iterating this approach leads to subsequently more accurate poses (as evident from the smaller lines caused by a more accurate overlay). The final pose estimate can also be verified visually. }
    \label{fig:overview}
\end{figure*}

\section{Introduction}
\label{sec:intro}
Visual localization is the problem of estimating the camera pose, \ie the position and orientation from which an image was taken, with respect to a known scene. 
Visual localization is a core component of many interesting applications such as self-driving cars~\cite{Heng2019ICRA} and other autonomous robots such as drones~\cite{Lim12CVPR}, as well as for augmented and virtual reality systems~\cite{Castle08ISWC,Lynen2015RSS}. 

Similar to other areas in computer vision, the availability of benchmark datasets such as \cite{Shotton13cvpr,Valentin20163DV,Kendall15iccv,Sattler2012bmvc,Sattler18cvpr,Badino_IV11,Maddern17ijrr} has served as a main driving force for research. 
Yet, there is a fundamental difference between visual localization and areas such as semantic segmentation and object detection in the way ground truth is obtained. 
For the latter, ground truth is provided by human annotations. 
However, humans are not able to directly predict highly accurate camera poses. 
Instead, ground truth is typically computed through a reference algorithm, \eg Structure-from-Motion (SfM). %
Thus, localization benchmarks do not measure absolute pose accuracy. 
Rather, they measure to what degree methods are able to replicate the results of the reference algorithm. 
Given that the reference approach itself will produce inaccuracies and errors in the pose estimates, we use the term ``reference poses" instead of ``ground truth poses". 

It is crucial that the reference algorithm generates poses with a higher accuracy than the actual localization methods evaluated on a benchmark. 
It is thus common to provide more data to the reference algorithm compared to what is made available to the localization approaches. 
For example, data from other sensors such as depth~\cite{Shotton13cvpr,Valentin20163DV},  Lidar~\cite{Maddern17ijrr}, an external motion capture system such as Vicon~\cite{Schoeps2019CVPR}, or additional images not available to the localization methods~\cite{Schoeps2019CVPR} can be used if available. 
This paper considers the case where only images are available. 
In this case, SfM is typically used as the reference algorithm, \ie the reference poses are obtained jointly from all test images whereas localization approaches typically localize a single image at a time. 
This should lead to more accurate reference poses compared to what can be obtained from a single image. 

In particular, we are interested in reference pose generation in the context of long-term localization, which is the problem of localizing images taken under different conditions, \eg day-night or seasonal changes, against a scene captured under a reference condition. 
Given that scenes change over time, long-term localization is an important problem in practice. 
The main challenge in this setting is data association, \ie establishing feature matches between images taken under different conditions. 
Naturally, this causes problems for generating reference poses using SfM algorithms, which themselves rely on local features such as SIFT~\cite{Lowe04ijcv} for data association. 
In previous work, we thus relied on human annotations to obtain feature matches between images taken under different conditions~\cite{Sattler18cvpr}. 
However, this approach is not scalable. 
Furthermore, human annotations of feature positions in images tend to be inaccurate, as they can easily be off by 5-10 pixels or more.

This paper is motivated by the observation that the reference poses for the nighttime test images of the Aachen Day-Night dataset~\cite{Sattler18cvpr,Sattler2012bmvc}, obtained from human annotations, are not accurate enough to benchmark state-of-the-art localization methods. 
This paper thus proposes a semi-automated approach to reference pose generation. 
Our method is inspired by previous work on pose verification via view synthesis~\cite{Taira18cvpr,Taira19iccv,Torii2018pami} and the observation that modern learned local features~\cite{Dusmanu19cvpr,Revaud19neurips} capture higher-level shape information. 
The latter allows feature matching between real images and 3D models, \eg obtained via multi-view stereo~\cite{Schoenberger2016ECCV}. 
As shown in Fig.~\ref{fig:overview}, our approach starts with a given initial pose estimate. 
It renders the 3D scene model from the current pose estimate. 
Feature matches between the actual and the re-rendered image are then used to refine the pose estimate. 
This procedure is repeated for a fixed number of iterations. 
Detailed experiments, for multiple ways to obtain initial poses, show that our approach yields more accurate pose estimates.

Re-rendering the image from its estimate pose enables visual inspection of the accuracy of the estimate. 
Using this aid, we observe that even larger differences in pose of 20cm or more can have little impact on the rendered image. 
This is not particularly surprising as the uncertainty of a pose estimate depends on the distance to the scene. 
However, it also implies that using fixed position and rotation thresholds on the pose error to measure localization accuracy~\cite{Shotton13cvpr,Sattler18cvpr} is not appropriate if there are significant changes in scene depth between test images. 
As a second contribution, we thus discuss and evaluate multiple evaluation measures that (explicitly or implicitly) use per-image uncertainty measures rather than global thresholds on pose errors.

In detail, this paper makes the following contributions: 
(\textbf{1}) we propose an approach based on view synthesis and learned features that can be used to generate reference pose for long-term visual localization benchmarks. 
(\textbf{2}) we provide a detailed experimental analysis of our approach, including studying different initialization approaches, different strategies for rendering and different features. 
(\textbf{3}) we show that the existing nighttime reference poses of the Aachen Day-Night dataset are not accurate enough to evaluate state-of-the-art long-term localization approaches. 
We further use our approach to obtain refined reference poses and show that current localization approaches achieve a much higher (up to $47\%$) pose accuracy than indicated by the original reference poses. 
(\textbf{4}) we extend the Aachen Day-Night dataset by additional nighttime test images, effectively doubling the number of available test images. 
We evaluate state-of-the-art localization approaches on the extended dataset and will provide a benchmark at \href{https://www.visuallocalization.net/}{visuallocalization.net}. 
(\textbf{5}) we discuss and experimentally study additional evaluation measures. 
(\textbf{6}) we will make source code for our approach and our evaluation measures publicly available to facilitate the creation of new benchmarks. 
(\textbf{7}) we provide a concise review of current trends in the area of visual localization.

\begin{figure*}[t]
    \centering
    \includegraphics[width=0.8\linewidth]{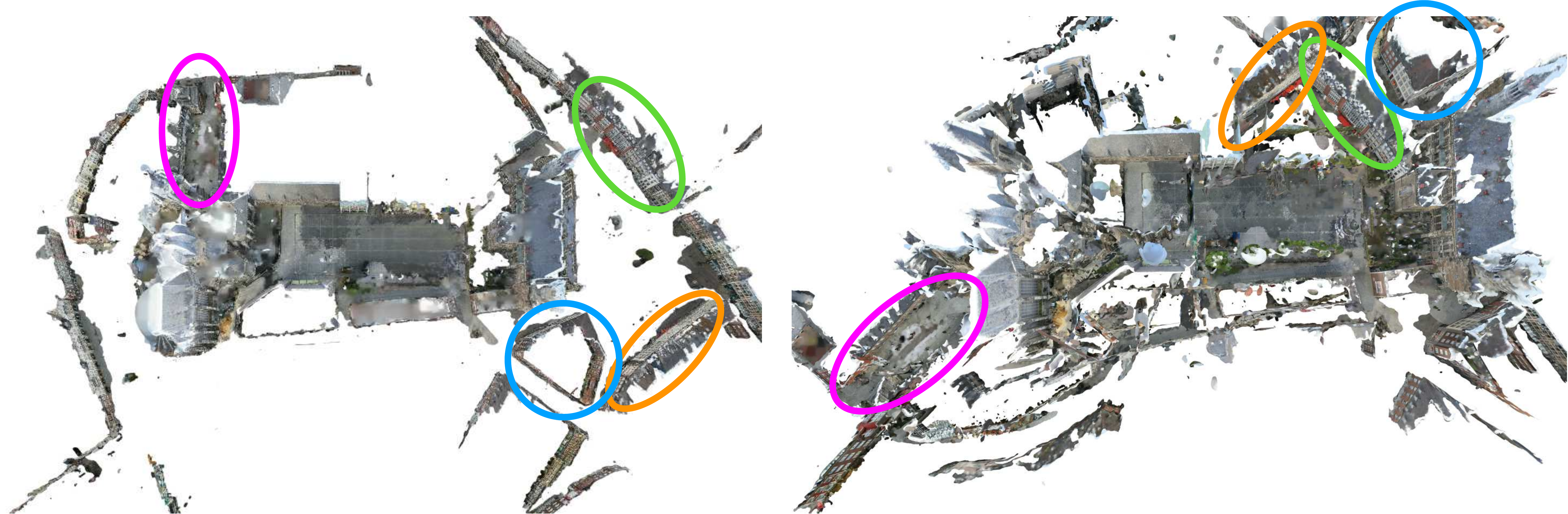}
    \caption{Multi-View Stereo reconstructions obtained from SfM models of the Aachen dataset using SIFT (left) and D2-Net (right) features (top-down view). D2-Net features are more robust to changes in conditions, \eg day-night and seasonal changes, than classic SIFT features, but also produce more false positive matches. This leads to connecting unrelated scene parts during the SfM process and ultimately in an incorrect 3D model. In contrast, SIFT correctly reconstructs the scene. Some wrong placements are illustrated through colored ellipses.}
    \label{fig:aachen_broken}
\end{figure*}

\section{Related work}
\label{sec:related_work}
Besides discussing related work on benchmark creation for visual localization and the use of view synthesis for pose estimation and verification, this section also aims at giving an interested reader a concise overview over main trends in the area of visual localization.

\PAR{Visual localization.} Traditionally, most visual localization algorithms have been based on a combination of local features and a 3D scene model~\cite{Se2002IROS,Robertson04BMVC,Li2010ECCV,Li2012ECCV,Choudhary12ECCV,Irschara09CVPR,Sattler2011ICCV,Jones11ijrr,Williams2007ICCV}. 
In most cases, the underlying 3D model is a sparse 3D point cloud constructed using SfM~\cite{Schoenberger2016cvpr,Snavely08IJCV} or SLAM~\cite{Davison07pami,MurArtal17tro}. 
Each point in this model has been triangulated from two or more local image features such as SIFT~\cite{Lowe04ijcv} or ORB~\cite{Rublee2011ICCV}. 
Thus, each 3D point can be associated with one or more local image descriptors. 
2D-3D correspondences between local features in a query image and 3D model points can be found using nearest neighbor search in the descriptor space. 
In turn, these 2D-3D matches can be used to estimate the camera pose of the query image by applying an $n$-point pose solver~\cite{Haralick94IJCV,Kukelova13ICCV,Kukelova2010ACCV,Larsson2017ICCV,Albl2016CVPR,Kneip11cvpr,Fischler81cacm} inside a hypothesize-and-verify framework such as RANSAC~\cite{Fischler81cacm} and its variants~\cite{Chum08PAMI,Lebeda2012BMVC,Raguram2013PAMI}.
Research on such 3D \textit{structure-based} methods has mostly focused on scalability, \eg by accelerating the 2D-3D matching stage~\cite{Li2010ECCV,Li2012ECCV,Choudhary12ECCV,Sattler2017PAMI,Donoser14CVPR,Lim12CVPR,Jones11ijrr,Cheng2019ICCV} 
and the use of image retrieval~\cite{Irschara09CVPR,Sattler2012bmvc,Sarlin2019CVPR,Taira18cvpr,Liu2017ICCV,Cao13CVPR}, 
by reducing memory requirements through model compression~\cite{Li2010ECCV,Cao2014CVPR,Camposeco2019CVPR,Lynen2015RSS,Dymczyk2015ICRA}, 
or by making the pose estimation stage more robust to the ambiguities encountered at scale~\cite{Li2012ECCV,Zeisl2015ICCV,Svarm2017PAMI,Larsson2016BMVC,Alcantarilla11ICRA,Aiger2019SoCG}.

Such approaches are computationally too complex for mobile devices with limited resources, \eg robots and smart phones. 
In order to achieve real-time localization on such devices, non-real-time global localization against a pre-built map is combined with real-time local camera pose tracking~\cite{MurArtal2017RAL,Middelberg2014ECCV,Lynen2015RSS,Schneider2018RAL,Kasyanov2017IROS,DuToit2017ICRA,Jones11ijrr,Ventura2014TVCG}. 
To this end, results from the localization process (most often 2D-3D inliers) are integrated into visual(-inertial) odometry or SLAM to prevent drift in the local pose estimates. 

Structure-based approaches rely on underlying 3D models, which are expensive to build at scale and costly to maintain~\cite{Sattler2017CVPR}. 
Alternatively, the absolute pose of a query image can be estimated from the relative poses to database images with known poses~\cite{Zhang06TDPVT,Zhou2020ICRA} or 2D-2D matches with multiple database images~\cite{Zheng2015ICCV}. It can also be estimated using local SfM models computed on the fly~\cite{Sattler2017CVPR}.

Instead of explicitly using an underlying 3D model, \textit{absolute pose regression} train a CNN to directly regress the camera pose from an input image~\cite{Brahmbhatt2018CVPR,Clark2017CVPR,Huang2019ICCV,Kendall15iccv,Kendall2017CVPR,Melekhov2017ICCVW,Naseer2017ICRA,Radwan2018RAL,Valada2018ICRA,Walch2017ICCV,Xue2019ICCV}. 
However, they are not consistently more accurate than simple image retrieval baselines~\cite{Arandjelovic2016CVPR,Torii2018pami,torii11} that approximate the pose of a query image by the poses of the top-retrieved database images~\cite{Sattler2019CVPR}. 
Furthermore, these approaches need to be trained specifically per scene. 
The latter problem can be overcome by \emph{relative pose regression} techniques~\cite{Balntas2018ECCV,Ding2019ICCV,Laskar2017ICCVW,Zhou2020ICRA,Saha2018BMVC}, which train CNNs to predict relative poses. 
In combination with image retrieval against a database of images with known poses, these relative poses can be used for visual localization. 
While recent work shows promising results~\cite{Ding2019ICCV,Saha2018BMVC,Zhou2020ICRA}, relative pose regression techniques do not yet achieve the same level of pose accuracy as methods explicitly based on 2D-3D matches. 

Rather than learning the full localization pipeline, \emph{scene coordinate regression} algorithms only replace the 2D-3D matching stage through a machine learning algorithm, typically either a random
forest~\cite{Shotton13cvpr,Cavallari2017CVPR,Cavallari20193DV,Cavallari2019TPAMI,Meng2017IROS,Meng2018IROS,Massiceti2017ICRA,Valentin2015CVPR,Valentin20163DV} or a
CNN~\cite{Brachmann2017CVPR,Brachmann2018CVPR,Brachmann2019ICCVa,brachmann2020ARXIV,Massiceti2017ICRA,Yang2019ICCV,Zhou2020CVPR}. 
For a given patch from an image, these methods predict the corresponding 3D point in the scene. 
The resulting in a set of 2D-3D matches can then be used for camera pose estimation. 
Scene coordinate regression techniques constitute the state-of-the-art in terms of pose accuracy in small scenes. 
However, they currently do not scale well to larger scenes. 
For example, ESAC~\cite{Brachmann2019ICCVa}, a state-of-the-art scene coordinate regression technique, localizes 42.6\% of all daytime query images of the Aachen Day-Night dataset~\cite{Sattler2012bmvc,Sattler18cvpr} within  errors of 25cm and 5$^\circ$. 
In contrast, SIFT-based Active Search~\cite{Sattler2017PAMI}, a classical structure-based method, localizes 85.3\% within the same error thresholds. 

\PAR{Learned local features.} 
State-of-the-art approaches for long-term localization~\cite{Dusmanu19cvpr,Sarlin2019CVPR,Germain20193DV,MLarsson2019ICCV,Stenborg2018ICRA,Yang2020ARXIV,Benbihi2019ICCV,Taira18cvpr,Taira19iccv} are based on local features and explicit 3D scene models.\footnote{See also  \href{https://www.visuallocalization.net/benchmark/}{visuallocalization.net/benchmark/}.} 
Classical handcrafted features such as ORB~\cite{Rublee2011ICCV}, SIFT~\cite{Lowe04ijcv}, and SURF~\cite{Bay2008CVIU} struggle to match features between images taken under strongly differing viewing conditions, \eg day and night or seasonal changes. 
Thus, long-term localization approaches typically use machine learning, both for image retrieval~\cite{Arandjelovic2016CVPR,Noh2017iccv,Radenovic2019PAMI} and for local features~\cite{Ono2018NIPS,DeTone2018CVPRWorkshops,Benbihi2019ICCV,Noh2017iccv,Yang2020ARXIV,Dusmanu19cvpr}.

Traditionally, local feature learning has focused on learning feature descriptors~\cite{Balntas2016BMVC,Brown2011PAMI,Ebel2019ICCV,Mishchuk2017NIPS,Simonyan2014PAMI,Simo2015ICCV,Tian2017CVPR,Tian2019CVPR}. 
However, it has been shown that the local feature detector often is the limiting factor~\cite{Taira18cvpr,Torii2018pami,Sattler18cvpr,Germain20193DV}. 
Thus, recent work trains feature detectors and descriptors jointly~\cite{Benbihi2019ICCV,DeTone2018CVPRWorkshops,Ono2018NIPS,Yang2020ARXIV,Wang2020ARXIV,Noh2017iccv}, 
leading to state-of-the-art feature matching performance for images taken under strongly differing conditions. 
Interestingly, using deeper layers of neural networks pre-trained on ImageNet~\cite{Deng2009CVPR} to define both  feature detector and descriptor leads to very competitive performance~\cite{Benbihi2019ICCV,Dusmanu19cvpr}. 
Equally important, such features are very robust to changes in different conditions, even though this might come at a price of more false positives (cf. Fig.~\ref{fig:aachen_broken}). 
We use this robustness to establish correspondences between real images and renderings of 3D models and the resulting 2D-3D matches to compute reference poses for benchmarking long-term visual localization. 
In addition, we benchmark state-of-the-art long-term localization approaches~\cite{Sarlin2019CVPR,Germain20193DV,Revaud19neurips,Dusmanu19cvpr} based on local features our reference poses.

\PAR{Semantic visual localization.} 
Besides using learned features that are more robust to changes in viewing conditions, long-term localization approaches also use semantic image segmentation~\cite{Budvytis2019BMVC,Garg2019IJRR,MLarsson2019ICCV,Stenborg2018ICRA,Schoenberger2018CVPR,Seymour2019BMVC,Shi2019ICIP,Taira19iccv,Toft2017ICCVW,Toft2018ECCV,Wang2019PAMI,Yu2018IROS}. 
These methods are based on the observation that the semantic meaning of scene elements, in contrast to their appearance, is invariant to changes. 
Semantic image segmentations are thus used as an invariant representation for image retrieval~\cite{Arandjelovic2014ACCV,Toft2017ICCVW,Yu2018IROS}, to verify 2D-3D matches~\cite{Budvytis2019BMVC,MLarsson2019ICCV,Stenborg2018ICRA,Toft2018ECCV} and camera pose estimates~\cite{Shi2019ICIP,Stenborg2018ICRA,Taira19iccv,Toft2018ECCV}, for learning local features~\cite{Garg2019IJRR,Schoenberger2018CVPR}, and as an additional input to learning-based localization approaches~\cite{Budvytis2019BMVC,Seymour2019BMVC,Wang2019PAMI}.

\PAR{View synthesis.} As shown in Fig.~\ref{fig:overview}, our approach iteratively renders a 3D model from a camera pose estimate and uses matches between the rendering and the actual image to refine the pose. 
Our approach takes inspiration from previous work on using view synthesis for pose estimation and verification. 
\cite{Sibbing133DV,Shan-3DV14} render detailed laser scans~\cite{Sibbing133DV} respectively dense Multi-View Stereo point clouds~\cite{Shan-3DV14} from new perspectives. 
They show that SIFT feature matching between the renderings and actual images is possible if both were taken from very similar poses. 
\cite{Torii2018pami} shows that view synthesis from very similar viewpoints (obtained from depth maps) improves SIFT feature matching between day and night images. 
\cite{Aubry-ACMTOG14} learns features that can be ma\-tch\-ed between paintings and renderings of a 3D model.
\cite{Valentin16threedv} first learns a randomized decision forest (RDF) and a hierarchical navigation graph using synthesized images (rendered from reconstructed scene models) and then uses the RDF and the graph for efficient and gradient-free localization of new query images.
In these works, view synthesis is used to create novel viewpoints in a given scene in order to enable camera pose estimation at all. 
In contrast, this paper focuses on using view synthesis to refine an initial pose estimate and to use it for generating reference poses for a long-term localization benchmark. 
Thus, the contributions of this paper center around a detailed experimental evaluation of the use of view synthesis to improve pose accuracy rather than on proposing a new method. 

\cite{Taira18cvpr,Taira19iccv} use view synthesis for automated pose verification. 
To this end, they render a dense laser scan point cloud from a set of given poses. 
They densely extract descriptors from each rendering and compare each descriptor against a descriptor extracted at the same pixel in the original image to compute an image-level similarity score. 
This score is then used to select the pose that best explains the input image. 
In contrast, this paper uses view synthesis to refine the camera pose estimates. 
While \cite{Taira18cvpr,Taira19iccv} automate pose estimation, their approach still has room for improvement, even if additional information such as semantics is used~\cite{Taira19iccv}. 
Thus, we use the rendering for visual inspection of the poses rather than automating the verification process.

\cite{Armagan17cvpr} utilizes synthesized views for improving an initial pose estimate with respect to a 2.5D street map (containing only the outline of the buildings).
Given a rendered view of the 2.5D map and a semantic segmentation of the image, they combine two networks and a line search strategy to compute a 3DoF pose correction (horizontal translation and yaw) to the initial pose.
Then the improved pose is used as the new input, and the correction procedure is applied iteratively.
This paper uses a similar strategy of iterative synthesis and correction but differs in several aspects.
\cite{Armagan17cvpr} focuses on the geolocalization in urban environment, while we aim at providing a general tool for creating accurate visual localization benchmarks.
This also results in the different choices of the input modality: we choose to use SfM models instead of 2.5D maps, which are specific to urban environments.
Moreover, our method is more generic in that it is able to correct the poses in 6DoF instead of 3DoF.

\PAR{Visual localization benchmarks.} 
This paper considers the visual localization problem, \ie the task of computing the full camera pose for a given image. 
Closely related is the visual place recognition problem of determining which place is visible in a given image, without necessarily estimating its camera pose. 
However, we will not discuss pure place recognition datasets that do not provide full 6DoF camera poses such as~\cite{Chen2017ICRA,Suenderhauf2015RSS,Torii2018pami,Torii15PAMI,milford2012seqslam}. 

Early localization benchmarks used SfM to reconstruct scenes from internet photo community collections such as Flickr. 
Query images were then obtained by removing some images from the reconstruction, together with all 3D points visible in only one of the remaining images~\cite{Li2010ECCV}. 
Examples for this approach to benchmark creation are the Dubrovnik, depicting the old city of Dubrovnik (Croatia), Rome~\cite{Li2010ECCV} and Landmarks 1k~\cite{Li2012ECCV} datasets. 
The latter two datasets consists of individual landmarks in Rome respectively around the world. 
The same approach was later also used for images taken under more controlled conditions, \eg the crowd-sourced Arts Quad~\cite{Crandall2011CVPR,Li2012ECCV} dataset, the scenes from the Cambridge Landmarks~\cite{Kendall15iccv} benchmark, and the San Francisco SF-0~\cite{Chen2011CVPR,Li2012ECCV,Sattler2017CVPR} dataset. 
Similarly, RGB-D SLAM algorithms~\cite{Newcombe2011ISMAR,Dai2017TOG} were used to obtain reference poses for the 7Scenes~\cite{Shotton13cvpr} and 12Scenes~\cite{Valentin20163DV} datasets. 
Both depict small indoor scenes captured with RGB-D sensors. 

Long-term localization benchmarks~\cite{Sattler18cvpr,CarlevarisBianco2016IJRR,SILDA} typically use images captured under a reference condition to represent the scene while images taken under different conditions are used as query. 
SLAM and SfM algorithms depend on data association between images. 
Thus, they tend to fail if images were taken under too dissimilar conditions. 
Using image sequences and / or multi-camera systems can allow using SLAM and SfM algorithms under stronger viewing condition changes. 
The former exploits the fact that it is not necessary to find matches between each query image and a reference image. 
Rather, finding enough matches for some query images is sufficient to register an entire sequence. 
The latter exploit the fact that a larger field-of-view typically leads to more matches. 
Both the SILDa~\cite{SILDA} and (extended) CMU Seasons~\cite{Badino_IV11,Sattler18cvpr} use sequences and multi-camera systems. 
SILDa depicts a single building block in London, UK under diferent conditions. 
The (extended) CMU Seasons dataset was constructed from images collected in and around Pittsburgh, US over the span of a year. 
For the (extended) CMU Seasons, additional humanly annotated matches were used in areas where cross-seasonal matching failed~\cite{Sattler18cvpr}. 
Human annotations were also used for the Mall~\cite{Sun2017CVPR} dataset to obtain initial pose estimates of test images with respect to a laser scan. 

Manually annotated matches are often not very precise~\cite{Sattler2017CVPR}. 
If available, additional sensors such as Lidar can be used to avoid the need for human annotations. 
The RobotCar Seasons~\cite{Maddern17ijrr,Sattler18cvpr}, depicting the city of Oxford, UK under various seasonal conditions, and the University of Michigan North Campus Long-Term Vision and LIDAR~\cite{CarlevarisBianco2016IJRR} datasets use Lidar data to obtain reference poses. 
However, human intervention might still be necessary if the scene geometry changes~\cite{Sattler18cvpr}. 

The Aachen Day-Night~\cite{Sattler2012bmvc,Sattler18cvpr} depicts the old inner city of Aachen, Germany. 
The 3D model of the scene was reconstructed from daytime images using SfM. 
Similarly, reference poses for daytime query images were also obtained using SfM. 
Since additional sensor data is not available and since SfM failed to provide reference poses~\cite{Sattler18cvpr}, manual annotations were used for a set of nighttime query images. 
To this end, a daytime image taken from a similar viewpoint was selected for each nighttime query. 
The pixel positions corresponding to 10 to 30 3D points visible in the daytime image were then annotated manually. 
\cite{Sattler18cvpr} estimated that the median mean position accuracy for the nighttime images is between 30cm and 40cm. 
However, in this paper, we show that the pose estimates are actually often worse. 
This observation motivates our approach for refining the original reference poses. 
We show that the refined poses are more accurate and are thus more suitable to measure the performance of state-of-the-art localization techniques. 
While this paper focuses on the Aachen Day-Night dataset, our approach is not specific to it and can be applied on other datasets as well.
As described in \Sec\ref{subsec:exp_setup}, it uses the same information as required for building SfM models, which is available in many visual localization benchmarks.
The scene models used in our approach, namely a SfM model and a dense mesh, can be generated using publicly available software packages (\eg COLMAP in our setup).

\section{Reference Pose Generation} %
\label{sec:verify_refine}
Typically, a visual localization dataset provides a set of images $\mathcal{I}: \{\img_i\}_{i=1}^{N}$ and the corresponding reference poses $\mathcal{T}: \{\T_i\}_{i=1}^{N}$ in a 3D model $\mathcal{M}$.
Our goal is to know whether the poses $\mathcal{T}$ are accurate (verification) and get more accurate reference poses if necessary (refinement).
Since each image in a visual localization dataset is usually treated individually, we consider a single image $\img$ and its (potentially inaccurate) pose $\T$ in this section.
$\T$ represents the camera pose with respect to the model $\mathcal{M}$.
More specifically, $\T$ is a $4\times4$ transformation matrix:
\begin{equation}
    \T = \begin{bmatrix}
    \R & \t \\
    \mathtt{0}_{1\times3} & 1
    \end{bmatrix},
\end{equation}
and $\pt = \R \cdot {}_{c}\pt + \t$ converts point coordinates in the camera frame ${}_{c}\pt$ to the coordinates in the model.

Given the 3D model $\mathcal{M}$, we first render a synthesized view $\img^{r}$ (or multiple rendered images) at pose $\T$ (\Sec\ref{subsec:rendering}).
Then learned features are extracted and matched between the actual image $\img$ and the synthesized image $\img^r$.
By analyzing the matched features, denoted as $\{\px_l\}\lrange{N_f}$ and $\{\px_l^r\}\lrange{N_f}$ for the actual and rendered images respectively, we can determine whether the pose $\T$ is accurate (\Sec\ref{subsec:matching_verification}).
Finally, we can back-project the 2D features from the rendered view $\{\px_l^r\}\lrange{N_f}$ to the 3D model $\mathcal{M}$ to get a set of 3D points $\{\pt_l^r\}\lrange{N_f}$.
From the 2D-3D correspondences $\{\px_l\}\lrange{N_f}$ and $\{\pt_l^r\}\lrange{N_f}$, we can calculate a more accurate pose $\T^{r}$ for the actual image (\Sec\ref{subsec:refining}).
The aforementioned process is repeated several times to get more accurate poses (cf. \Fig\ref{fig:overview}).
We also discuss different methods to quantify the uncertainties of the resulting poses (\Sec\ref{subsec:uncertainty_quantification}), which are useful for defining localization accuracy metrics (cf. \Sec\ref{sec:loc_metrics:direct}).

For simplicity of presentation, we assume that all the 2D features $\{\px_l^r\}\lrange{N_f}$ have a valid back-projection in $\mathcal{M}$ and all the 3D points $\{\pt_l^r\}\lrange{N_f}$ are inliers in the refinement process.
In practice, we remove 2D features with invalid depth (\eg due to an incomplete the model $\mathcal{M}$) and reject outliers using LO-RANSAC~\cite{Lebeda2012BMVC}. 
For simplicity, we assume that the features are ordered based on matches: 
for a feature $\px_l$ in the real image, the corresponding matching feature in a rendering is $\px_l^r$.

\subsection{Rendering Synthesized Views}
\label{subsec:rendering}

There are different methods to render synthesized views from a pose $\T$ with respect to a scene model $\mathcal{M}$.
In this work, we investigate view synthesis from two different scene models: a 3D point cloud with SIFT descriptors and a 3D mesh.
In the process of generating reference poses using SfM, the scene is typically reconstructed as a 3D point cloud, where each point is associated with a descriptor, \eg SIFT.
A 3D mesh can be further generated using Multi-View Stereo.
Therefore, these two models are readily available from the standard process for generating reference poses.

To render images from a 3D mesh, there are various off-the-shelf renderers that can be used. 
As for a point cloud with descriptors, we follow \cite{Pittaluga19cvpr} and train a CNN to reconstruct the images from such a scene representation.
The network uses a U-Net architecture \cite{Ronneberger15icmicci}.
The input to the network is a 3D tensor of size $h \times w \times 129$, where $h$ and $w$ are the height and width of the image to be synthesized.
The $129$ channels consists of a depth channel and one channel per byte in the SIFT descriptor ($128$ bytes).
The input is constructed by finding and projecting the visible points in the point cloud to the pose to render and then filling the input tensor at the pixel coordinates where there is a projected 3D point.
The output of the network is the synthesized image at a given pose.
For details of the method (\eg training and evaluation), we refer the reader to \cite{Pittaluga19cvpr}.
While each rendering technique alone is sufficient in certain cases, combining the two rendering methods utilizes the information from different scene models and results in the best performance in our experiment (cf. \Sec\ref{subsec:exp_ablation}).

\subsection{Matching Features with Synthesized views}
\label{subsec:matching_verification}
To extract and match features between the real images $\img$ and the rendered images $\img_r$, we choose to use learning-based local features.
This is due to the fact that the rendered images usually have large appearance change compared with the real night images.
Traditional features, such as SIFT, rely on low level image statistics and  are not robust to day-night condition change and rendering artifacts.
In particular, we choose to use the D2-Net feature \cite{Dusmanu19cvpr} in our pipeline, which uses a single CNN for joint feature detection and description and achieves state-of-the-art matching performance in challenging conditions.

For the images rendered using the two rendering techniques, we extract and match features between each rendered image and the real image individually.
We then directly aggregate the feature matches obtained from both rendered images for the next step.
Note that after obtaining the 2D feature matches, we can already verify whether there exists pose errors in the reference poses by checking the matching locations in the rendered and real images (cf. \Fig\ref{fig:project_old_ref_poses} and \Fig\ref{fig:d2net_fail} for large and small pose errors respectively): 
if the real and rendered images are taken from the same pose, the two features $\px_l$ and $\px_l^r$ should be found at identical 2D positions (up to noise in the feature detection stage). 
Similarly, a large 2D distance $||\px_l - \px_l^r||_2$ is indicative for a significant difference in pose.

\subsection{Refining Reference Poses}
\label{subsec:refining}
Given $N_f$ matched features $\{\px_l\}\lrange{N_f}$ and $\{\px_l^r\}\lrange{N_f}$ between the real and rendered images, we first back-project the features in the rendered images to $\mathcal{M}$ to get the corresponding 3D points as  $\{\pt_l^r\}\lrange{N_f}$
\begin{equation}
    \pt^r_l = \pi^{-1}(\px_l^r, \T, K, D, \mathcal{M}),
\end{equation}
where $\pi:\pt \rightarrow \px $ is the camera projection function and $\pi^{-1}$ the inverse. $K$ and $D$ are the intrinsics and distortion parameters respectively.
In practice, we get the depth map at $\T$ in the process of rendering images from the 3D mesh, and the depth at $\px_l$ can be directly read from the depth map.
After finding the 3D points, the refined reference pose $\T^r$ can be computed by solving a nonlinear least-squares problem
\begin{equation}
    \T^r = \underset{\T}{\argmin}
    \sum\lrange{N_f} \lVert \pi(\pt^r_l, \T, K, D) - \px_l \rVert^2.
    \label{eq:pose_refine}
\end{equation}
We minimize \eqref{eq:pose_refine} over the inliers of a pose obtained by LO-RANSAC.
Note that it is possible to additionally refine $K$ and $D$ in the above optimization.
This could help correct errors in the intrinsics and distortion parameters but potentially make the optimization problem less stable (\eg when there are few feature matches).
In addition, relatively accurate $K$ and $D$ values are required for rendering reasonable synthesized views for successful feature matching, regardless of whe\-ther they are refined in the optimization.

\subsection{Uncertainty Quantification}
\label{subsec:uncertainty_quantification}
To use the refined pose $\T^r$ for evaluating localization accuracy, it is also important to quantify the uncertainty of the refined pose for meaningful interpretation of the difference between the pose to evaluate and the reference pose.
For example, if two poses to be evaluated are both within the uncertainty range of the reference pose, they should be considered equally accurate, regardless of their absolute differences with respect to the reference pose.
It is thus a typical practice to consider the uncertainty of the reference poses in the accuracy evaluation of visual localization methods (cf. \Sec\ref{sec:loc_metrics:direct}).
Next, we introduce two commonly used methods for covariance estimation \cite[Ch.~5]{Hartley03book} and a sampling strategy explored in this paper.

\PAR{First Order Approximation}
For the nonlinear least squares problem \eqref{eq:pose_refine}, the covariance of $\T^r$ can be computed by
$
    \mathbf{\Sigma}_{\T^r} =
    (\sum_{l=1}^{N_f}
    \J^\top_{l} \mathbf{\Sigma}^{-1}_{\px} \J_{l})^{-1},
    \label{eq:cov_first_order}
$
where $\J_{l} = \partial \px_l / \partial \T$ is the Jacobian (evaluated at $\T^r$) of the $l$th landmark observation with respect to the camera pose
\footnote{
In solving \eqref{eq:pose_refine}, iterative methods (\eg Gauss-Newton) are usually used, where $\J_{l}$ is actually the Jacobian with respect to a small perturbation around the current estimate.
},
and $\mathbf{\Sigma}_{\px}$ is the covariance of the landmark observation $\px$.
$\mathbf{\Sigma}_{\px}$ is usually assumed to be a diagonal matrix and the same for each observation.
Then the uncertainty of the camera position and rotation, denoted as $\delta^c$ and $\delta^r$, can be calculated from positions and rotations sampled from a multivariate Gaussian distribution $\mathcal{N}(\mathbf{0}, \mathbf{\Sigma}_{\T^r})$:
\begin{equation}
    \delta^c = \text{median}(\{\vert\delta\mathbf{c}_{n}\vert\}_{n=1}^{N_s})
    \quad \delta^r = \text{median}(\{\vert\delta\mathbf{r}_{n}\vert\}_{n=1}^{N_s}),
    \label{eq:uncertainty_first_order}
\end{equation}
where $\vert\cdot\vert$ denotes 2-norm. $\delta\mathbf{c}_n$ and $\delta\mathbf{r}_n$ are the sampled position and rotation respectively. We slightly abuse the notation and denote the number of samples as $N_s$ in this Section.
\PAR{Monte Carlo Estimation}
An alternative method is to use Monte Carlo method. In particular, from the 3D points $\{\pt^r_l\}\lrange{N_f}$ and refined pose $\T^r$, we first calculate the ideal (\ie no noise) observations $\{\bar\px_l\}\lrange{N_f}$, where $\bar\px_l = \pi(\pt^r_l, \T^r, K, D)$.
A set of noisy measurements $\{\tilde{\px}_l\}\lrange{N_f}$ can be simulated by $\tilde{\px}_l = \bar{\px}_l + \mathbf{n}_{\px}$, where $\mathbf{n}_{\px}\in\mathcal{N}(\mathbf{0}, \mathbf{\Sigma}_{\px})$.
We then can compute a camera pose estimate $\T^m$ using the method in \Sec\ref{subsec:refining} with the simulated feature locations $\{\tilde\px_l\}\lrange{N_f}$ instead of the actual ones $\{\px_l\}\lrange{N_f}$.
After repeating the above process $N_s$ times, the uncertainty of camera position $\sigma^{c}_m$ and rotation $\sigma^{r}_m$ can be computed as
\begin{equation}
    \delta^{c}_{m}= \text{median}(\{\epsilon^{c}_n\}_{n=1}^{N_s})
    \quad \delta^{r}_{m} = \text{median}(\{\epsilon^{r}_n\}_{n=1}^{N_s}),
    \label{eq:mc_uncertainties}
\end{equation}
where $\epsilon^c_n, \epsilon^r_n = \T^r \boxminus \T^m_n$ (cf. \eqref{eq:pose_diff}).

\PAR{Sampling Uncertainty}
We also explore a sampling strategy for uncertainty estimation in this paper.
In particular, for a sampling ratio $k$ (\eg $50\%$), we first randomly sample from all the 2D-3D matches (\ie all the inliers used in \eqref{eq:pose_refine}).
We then apply LO-RANSAC to the sampled subset and solve the nonlinear optimization problem \eqref{eq:pose_refine} using the inliers returned by LO-RANSAC to get a pose $\T^s$.
The sampling and solving process is repeated multiple times (typically $50$ times in our experiment), resulting in multiple pose estimates $\{\T^s_n\}_{n=1}^{N_s}$.
The sampling uncertainty for the camera position $s^{c}_{k}$ and rotation $s^{r}_{k}$ are calculated as
\begin{equation}
    s^{c}_k = \text{median}(\{\epsilon^{t}_n\}_{n=1}^{N_s})
    \quad s^{r}_k = \text{median}(\{\epsilon^{r}_n\}_{n=1}^{N_s}),
    \label{eq:sampling_uncertainties}
\end{equation}
where $\epsilon^c_n, \epsilon^r_n = \T^r \boxminus \T^s_n$ (cf. \eqref{eq:pose_diff}).
We calculate the sampling uncertainties for different sampling ratios.

We would like to highlight the difference between the uncertainties computed from the above methods and the absolute uncertainties.
The absolute uncertainties reflect the differences between the refined poses and the \emph{unknown} ground truth, which cannot be calculated directly.
The above uncertainties, on the other hand, evaluate the variance (by approximation or computing statistics from randomized sampling) with respect to the refined poses, which are essentially the local minima in the optimization problem \eqref{eq:pose_refine}.
Therefore, these uncertainties tend to be smaller than the actual uncertainties, since the local minima can hardly be the actual ground truth poses.
We will further discuss how to consider these uncertainties in the context of evaluation in \Sec\ref{sec:loc_metrics:direct}.

\subsection{Discussion}
The method proposed in this section essentially estimates more accurate poses from some potentially inaccurate initial estimates.
Yet, it can not only be used to verify and refine existing reference poses, but also to easily extend existing visual localization datasets.
For example, to add more images to an existing localization dataset, one only needs to provide coarse initial poses for these images, which can be obtained by, for example, manually selecting the most similar images.
This is useful especially for images with large appearance difference compared with the localization database (e.g., adding nighttime images to a localization database constructed from daytime images), where accurate poses cannot be reliably estimated using SfM directly.

\section{Metrics for Localization Accuracy}
\label{sec:loc_metrics}

The reference poses generated using SfM or our method are inherently subject to inaccuracies, which complicates the evaluation process.
For example, the difference between the reference pose and a pose to evaluate is no longer a meaningful metric if the actual error (\ie the difference between the pose to evaluate and the unknown ground truth) is comparable to the uncertainty in the reference pose.
Therefore, it is a common practice to set certain thresholds for the reference poses based on their uncertainties, and measure whether the poses to evaluate lie within those thresholds.
Unfortunately, quantifying the uncertainties in the reference poses is a highly non-trivial task in itself. %
The actual uncertainties depend on various factors, such as the depth of the scene and the accuracy of the local features.
In this section, we first discuss several performance metrics based on directly considering the uncertainties in pose space. 
We then discuss a performance metric based on the re-projection of the scene points, which removes the necessity of directly quantifying the pose uncertainty.

\subsection{Direct Pose Uncertainty-Based Measures}
\label{sec:loc_metrics:direct}
Direct pose uncertainty-based measures analyze the position and rotation error between the reference and estimated poses.
Typically, given a reference pose $\T$ and a pose to evaluate $\hat{\T}$, the position and rotation error $\epsilon^t, \epsilon^r = \T \boxminus \hat{\T}$ are computed as \cite{Sattler18cvpr}:
\begin{equation}
    \epsilon^t = \lVert \t -\hat{\t} \rVert_2, \quad
    \epsilon^r = \arccos(\frac{1}{2}(\text{trace}(\R^{-1}\hat{\R}) - 1)).
    \label{eq:pose_diff}
\end{equation}
To account for the uncertainties in the reference poses, we can either use a set of fixed thresholds for all the images in a dataset or define thresholds for each image individually.

\PAR{Fixed error thresholds.}
We can define a set of $N_e$ increasing error thresholds $E^{\text{fixed}} = \{\mathbf{e}^{\text{pose}}_j\}_{j=1}^{N_e}$, where $\mathbf{e}_j = (c_j, r_j)$ contains both position and orientation thresholds.
These thresholds apply to all the images in a dataset.
A pose is said to be below a threshold $\mathbf{e}_j$ if $\epsilon^c < c_j$ and $\epsilon^r < r_j$.
The overall localization accuracy is the percentages of images that are localized within these thresholds, and higher values indicate better performance.
For example, the error thresholds for Aachen night time images on \href{https://www.visuallocalization.net/}{visuallocalization.net} are 0.5/1.0/5.0 \si{m} and 2.0/5.0/10.0 \si{deg}
\footnote{Note that these thresholds were used \emph{before} we updated the reference poses with the poses generated using our method. A set of tighter thresholds, as in \Tab\ref{tab:eval_original_images} and \ref{tab:eval_overall}, are used now since the new reference poses are more accurate.}
,
and the localization accuracy is reported as three percentages corresponding to the these categories.

\PAR{Per image error thresholds}
Using the same thresholds for all the images in a dataset, however, has limitations.
The uncertainties are image-dependent if, as in our case, the poses are calculated by minimizing the reprojection errors of 2D-3D correspondences.
The position uncertainty is lower for images observing landmarks that are closer to the camera.
Ideally, these uncertainties should be taken into consideration to choose the error thresholds \emph{per image}.
As shown in \Sec\ref{subsec:uncertainty_quantification}, there are different ways of computing the pose uncertainty for each image, which can be used as per image error thresholds.
For the first order and Monte Carlo uncertainties, we can simply use \eqref{eq:uncertainty_first_order} and \eqref{eq:mc_uncertainties} as thresholds.
In terms of sampling uncertainties, we can choose a set of sampling uncertainties $E^{\text{sample}}_{i} = \{\mathbf{s}_{k}\}_{k=k_1, k_2, ...}$ as error thresholds, where $\mathbf{s}_k = \{s_k^{c}, s_k^{r}\}$ is the sampling uncertainty with sampling ratio $k$.
For example, in our experiment, we use a set of thresholds calculated from sampling ratios of $50\%$, $30\%$ and $10\%$ respectively.
However, as discussed before, the uncertainties in \Sec\ref{subsec:uncertainty_quantification} tend to be lower than the (unknown) absolute uncertainties. 
Therefore, using these uncertainties as error thresholds tends to under-estimate the accuracy of localization algorithms (cf. \Sec\ref{subsec:exp_stoa_diff_metrics}).

\subsection{Indirect Pose Uncertainty-Based Measures}
\label{sec:loc_metrics:indirect}
To avoid the need to consider the uncertainties in 6 DoF poses (which is non-trivial as seen before), we follow the literature on object pose estimation and measure pose accuracy based on reprojections~\cite{Hinterstoisser2012ACCV}. 
More precisely, we measure the difference between the reprojection of a set of 3D points in the reference and estimated poses.
Intuitively, perturbations to the camera pose will result in the changes of the reprojected 2D locations of 3D points.
Therefore, we can define certain thresholds around the reprojection of the 3D points as an \emph{indirect} measure of the pose uncertainty.
A key advantage of this approach is that the error thresholds can be defined on the image plane. 
While we use the same thresholds for all the images, this actually results in per-image uncertainty thresholds in pose space: 
the same change in reprojection error will typically result in a  position error that increases with increasing distance of the camera to the scene. 
Formally, we define the following metric:

\PAR{Maximum reprojection difference.}
The maximum distance between the projected points in the reference pose $\T^r_i$ and the estimated pose $\hat{\T}_i$ is used to measure the localization error:
\begin{equation}
    r_i^{\infty} = \underset{l\in [1, N_f^i]}{\max} \lVert \pi(\pt_l^r, \T_i^r) - \pi(\pt_l^r, \hat{\T}_i) \rVert_2,
\end{equation}
where the intrinsics and distortion parameters are omitted for simplicity.
Similar to  the pose error, a set of reprojection thresholds $E^{\text{rep}}=\{e^{\text{rep}}_j\}\jrange{N_e}$ are selected, and the percentages of the images with $r_i^{\infty}$ lower than these thresholds are used to indicate the overall accuracy on the dataset.
We slightly abuse $N_e$ here to denote the number of error thresholds in general.

\begin{figure*}[t]
    \centering
    \includegraphics[width=0.85\linewidth]{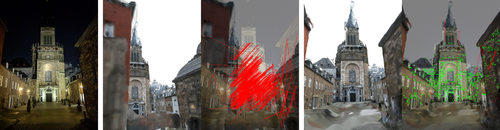}
    \includegraphics[width=0.85\linewidth]{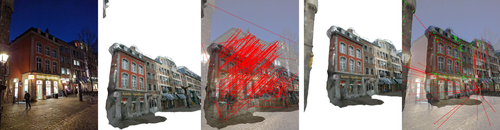}
    \includegraphics[width=0.85\linewidth]{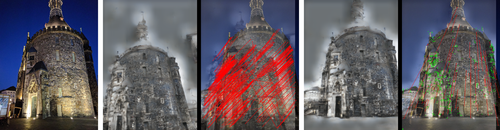}
    \includegraphics[width=0.85\linewidth]{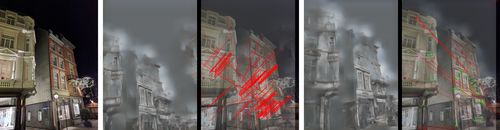}
    \caption{Comparison of images rendered from the original and refined (ours) reference poses of the nighttime images in Aachen Day-Night dataset.
    \textbf{First column}: nighttime images; \textbf{Second column}: images rendered from the existing reference poses, overlay of the rendering and the image together with D2-Net matches between the two; \textbf{Third column}: images rendered from our refined poses and the corresponding overlays with D2-Net matches.
    The top two rows render a Multi-View Stereo (MVS) mesh and the bottom two use Structure-from-Motion inversion~\cite{Pittaluga19cvpr} (invSfM).
    The colored lines visualize D2-Net feature matches.  Green is used to indicate that the 2D location difference between a feature in the real image and its match in the rendered image is below $20$ pixel.} %
    \label{fig:project_old_ref_poses}
\end{figure*}

\section{Experimental Evaluation}
\label{sec:experiments}
To demonstrate the value of the proposed method, we first use our method to analyze the reference poses of the nighttime query images in the Aachen Day-Night dataset (\Sec\ref{subsec:exp_analyzing_aachen}).
Then, we extend the dataset with new nighttime query images and generate the corresponding reference poses using our method (\Sec\ref{subsec:exp_extending_aachen}). We also compare our method against baseline methods of directly matching features (SIFT and D2-Net) and computing poses via SfM models.
To understand the impact of the different parameters in our method, we perform an extensive ablation study regarding different learned features, different rendering techniques, and the stability of our reference poses (\Sec\ref{subsec:exp_ablation}).
Finally, we evaluate state-of-the-art localization methods on both the original and the extended Aachen Day-Night datasets based on the performance metrics discussed in \Sec\ref{sec:loc_metrics} (\Sec\ref{subsec:exp_stoa_diff_metrics}).

In this paper, we focus on the Aachen Day-Night dataset~\cite{Sattler2018cvpr,Sattler2012bmvc}. 
This is motivated by our observation that the reference poses for the nighttime images are the least accurate reference poses among the three datasets from~\cite{Sattler2018cvpr}. 
At the same time, the dataset is becoming increasingly popular in the community, \eg \cite{Sarlin2019CVPR,Yang2020ARXIV,Wang2020ARXIV,Benbihi2019ICCV,Dusmanu19cvpr,Brachmann2019ICCVa,Mishchuk2017NIPS,Shi2019ICIP,Cheng2019ICCV,Revaud19neurips,Germain20193DV,Sarlin2020CVPR,Zhang2019ICCP} have already been evaluated on the dataset. 
However, our approach is generally applicable and can be applied to other datasets as well. 
Note that we only consider the nighttime query images in this paper as SfM already provides accurate reference poses for the daytime queries of the Aachen Day-Night dataset.
In contrast, the authors of the dataset reported in \cite{Sattler18cvpr} that localizing the nighttime images directly against the SfM model built from daytime images resulted in highly inaccurate poses due to a lack of sufficiently many SIFT feature matches.
This is due to the strong illumination changes between the nighttime and daytime images and the limited repeatability of the SIFT descriptor to such strong changes \cite{Zhou16eccvw}.
With insufficient feature matches, pose estimation using SIFT would either fail or have erroneous results due to ill-conditioned configurations (\eg the feature matches may concentrate in a small, well illuminated region in the image, resulting in an inaccurate pose estimate).
We observe similar failure cases when using SIFT to add new nighttime images to the Aachen Day-Night dataset (see \Fig\ref{fig:extension_sift_fail} and the discussion in \Sec\ref{subsec:exp_extending_aachen} for examples).

\subsection{Experimental Setup and Data Acquisition}
\vspace{-1ex}
\label{subsec:exp_setup}
\PAR{Additional data capture.} To extend the Aachen Day-Night dataset, we captured another 119 nighttime images and 119 daytime images with the camera of a Nexus 5X smart phone in July 2017. 
The nighttime and daytime images form pairs of photos taken from very similar poses. 
Registering the daytime images against the reference SfM model provided by the Aachen Day-Night dataset then yields initial pose estimates for the new nighttime queries.
Both the original and the newly captured nighttime images have a resolution of $1600\times1200$ pixels (the diagonal is thus of $2000$ pixels).

\PAR{Scene model generation.} 
Our approach to refine camera poses requires an underlying 3D scene model. 
The Aachen Day-Night dataset provides a \emph{reference SfM model} consisting of 4,328 database images and 1.65M 3D points triangulated from 10.55M SIFT features~\cite{Sattler2018cvpr}. 
This publicly available reference model is a sub-model of a larger \emph{base SfM model} that was reconstructed using COLMAP~\cite{Schoenberger2016cvpr}. 
This base model also contains images from a set of videos as well as the daytime queries, resulting in a SfM model with 7,604 images and 2.43M 3D points triangulated from 17.75M features. 
This model was registered against the original Aachen SfM model from~\cite{Sattler2012bmvc} to recover the scale of the scene. 
The reference model was obtained by removing the sequences and query images from the base model. 

We started from the base model and created an \emph{extended SfM model}. 
We registered the additional daytime images and an additional image sequence\footnote{Using one of the original videos and extracting images at a higher frame rate.} against the base model while keeping the poses of the base model images fixed. 
The resulting model contains 12,916 images and 3,90M 3D points triangulated from 32.19M SIFT features. 
We used this extended base model when creating our new reference poses. 

We removed all query images and the newly added sequence images from the extended base model to create an \emph{extended reference SfM model} consisting of 6,697 images and 2.32M points triangulated from 15.93M SIFT features.
This model will be used to benchmark localization algorithms on our extended Aachen Day-Night dataset. 
We will make this new reference model publicly available, but will withhold the base models and the reference poses for the query images. 
Instead, we will provide an evaluation service on \href{https://www.visuallocalization.net/}{visuallocalization.net}. 
The motivation behind publishing this smaller dataset is to make sure that the reference poses were computed from additional data not available to localization algorithms. 
The inclusion of the original sequences is necessary as some of the newly added nighttime queries depict places not covered in the original reference model. 

In addition to the extended models, we also created a colored 3D mesh of the scene. 
We used COLMAP's Multi-View Stereo pipeline~\cite{Schoenberger2016ECCV} to obtain a dense point cloud. 
Screened Poisson surface reconstruction~\cite{Kazhdan2013TOG} of the point cloud then yields a colored mesh.

\PAR{Rendering.} 
Our method requires rendering the scene from estimated poses. 
For each pose, we generate two renderings: 
(1) we render the MVS \emph{mesh}, 
(2) we use the SfM inversion approach (\emph{invSfM}) from \cite{Pittaluga19cvpr} to recover an image directly from a rendering of the extended base model. 
We use our own implementation of invSfM. 
Note that we only use the CoarseNet stage and skip the VisibNet and RefineNet. 
We use the MVS mesh to determine which points are visible instead of VisibNet.
While skipping RefineNet reduces image quality, we found the results to be of sufficient quality. 
Moreover, as shown in \cite{Pittaluga19cvpr}, RefineNet mostly improves the color of the rendered image with respect to the CoarseNet.
Since D2-Net feature used in our method is quite robust to such changes in the view condition, we do not expect skipping RefineNet would have a large impact on the performance of our method.
Fig.~\ref{fig:project_old_ref_poses} shows example renderings obtained from the mesh and invSfM.

\PAR{Implementation details.} If not mentioned otherwise, 
we extract D2-Net features~\cite{Dusmanu19cvpr} from both rendered images. %
The refinement process is repeated for 5 iterations.
We use single scale features since the initial pose estimates are accurate enough such that multi-scale processing is not required. %
To determine whether our refinement succeeded, we only accept the refined pose when there are more than 10 effective inliers\footnote{The effective inlier count takes the spatial distribution of the matches in the image into account. It has been shown to be a better measure than the raw inlier count~\cite{Irschara09CVPR}.} found by LO-RANSAC~\cite{Lebeda2012BMVC,Sattler2019Github} from the input 2D-3D matches, using the P3P solver from~\cite{Kneip11cvpr}. 
More precisely, we subdivide each image into a $50\times50$ grid and count at most one inlier per cell. The cell size and the inlier threshold are determined experimentally.

\begin{figure}[t]
    \centering
    \includegraphics[width=0.49\linewidth]{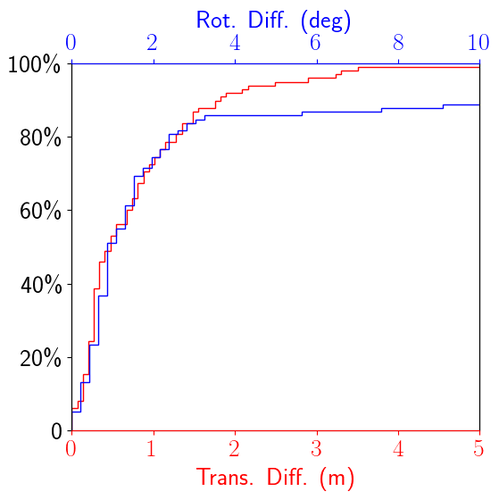}
    \includegraphics[width=0.49\linewidth]{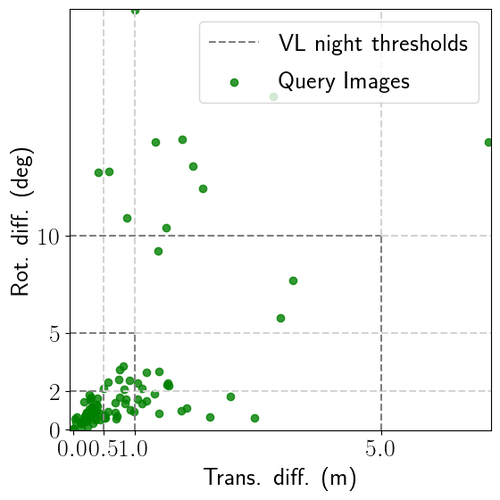}
    \caption{Differences between the original reference poses and the refined reference poses (our method). \textbf{Left}: Cumulative distribution of position and rotation differences. \textbf{Right}: Distribution of the position and rotational differences. The position and rotation thresholds (0.5/1.0/5.0 \si{m}, 2/5/10 \si{deg}) used in~\cite{Sattler2018cvpr} and  \href{https://www.visuallocalization.net/}{visuallocalization.net} (VL) are also shown for reference.}
    \label{fig:refined_gt_pose_change}
\end{figure}

\begin{table*}[t]
    \caption{Evaluation of state-of-the-art localization methods in the original Aachen nighttime images. We evaluate results submitted by the authors to \href{https://www.visuallocalization.net/}{visuallocalization.net} on both the original and our refined poses. We compare the methods based on the Pose Error, \ie the percentage of queries localized within fixed error thresholds of the reference poses. As can be seen, our more accurate reference poses yield a better measure of pose accuracy. For our poses, we also report results for two additional metric: the percentage of queries localized within sampling-based thresholds (Sampling) of the reference poses (cf. \Sec\ref{sec:loc_metrics:direct}) and the percentage of queries with maximum reprojection errors within given error thresholds in pixels  (Reprojection Diff.) (cf. \Sec\ref{sec:loc_metrics:indirect}). }
    \label{tab:eval_original_images}
    \centering
    \setlength{\tabcolsep}{5pt}
    \begin{tabular}{ l | c | c | c | c}
        \toprule
         & \textbf{Original Poses} & \multicolumn{3}{c}{\textbf{Refined Poses}} \\
         & \multicolumn{1}{c|}{\begin{tabular}{@{}c@{}}Pose Error\\ \scriptsize{0.25\si{m},2$^\circ$/0.5\si{m},5$^\circ$/5\si{m},10$^\circ$}\end{tabular}}
         & \multicolumn{1}{c}{\begin{tabular}{@{}c@{}}Pose Error\\ \scriptsize{0.25\si{m},2$^\circ$/0.5\si{m},5$^\circ$/5\si{m},10$^\circ$}\end{tabular}}
         & \multicolumn{1}{c}{\begin{tabular}{@{}c@{}}Sampling \\ \scriptsize{(50\%/30\%/10\%)} \end{tabular}} 
         & \begin{tabular}{@{}c@{}}Reprojection Diff. \\ \scriptsize{(10/20/50/100 px)} \end{tabular} \\
         \midrule
         Active Search v1.1~\cite{Sattler2018cvpr} 
         & 13.3/27.6/56.1  
         & 39.8/49.0/64.3 
         & 2.0/4.1/11.2
         & 28.6/39.8/52.0/62.2 \\
         D2-Net~\cite{Dusmanu19cvpr} 
         & 18.4/45.9/88.8  
         & 74.5/86.7/100.0 
         & 7.1/13.3/35.7
         & 46.9/68.4/89.8/98.0\\
         DELF~\cite{Noh2017iccv} 
         & 14.3/40.8/85.7 
         & 54.1/75.5/96.9 
         & 4.1/5.1/14.3
         & 28.6/56.1/78.6/88.8\\
         DenseVLAD~\cite{Torii2018pami} + D2-Net~\cite{Dusmanu19cvpr} 
         & 14.3/39.8/74.5 
         & 66.3/75.5/84.7 
         & 7.1/8.2/24.5
         & 45.9/65.3/77.6/82.7\\
         Hierarchical Localization~\cite{Sarlin2019CVPR} 
         & 17.3/43.9/76.5 
         & 68.4/77.6/88.8 
         & 7.1/9.2/24.5
         & 41.8/65.3/78.6/85.7  \\
         NetVLAD~\cite{Arandjelovic2016CVPR} + D2-Net~\cite{Dusmanu2019cvpr}
         & 17.3/43.9/85.7 
         & 84.7/90.8/96.9 
         & 8.2/11.2/40.8
         & 51.0/75.5/92.9/95.9 \\
         R2D2 V2 20K~\cite{Revaud19neurips} 
         & 19.4/48.0/88.8 
         & 75.5/90.8/100.0 
         & 8.2/14.3/36.7
         & 51.0/70.4/92.9/95.9 \\
         \bottomrule
    \end{tabular}
\end{table*}

\subsection{Refining the Original Aachen Nighttime Poses}
\label{subsec:exp_analyzing_aachen}
In a first experiment, we analyze the accuracy of the reference poses for the 98 original nighttime queries of the Aachen Day-Night dataset. 
We show that the original reference poses are inaccurate and that our refinement approach considerably improves the pose accuracy.

Our approach used the original poses for initialization. %
For $3$ out of the $98$ images, our method failed to find sufficiently many 2D-3D matches, mostly due to an incomplete mesh (see \Sec\ref{subsec:exp_ablation}).
For the failure cases, we simply kept the existing reference poses.

\PAR{Qualitative evaluation.} 
\Fig\ref{fig:project_old_ref_poses} visually compares the original reference poses with our refined poses.
As can be seen, the existing reference poses, obtained from manual annotated 2D-3D matches, can be rather inaccurate. %
In contrast, our method generates reference poses such that the rendering from the refined pose is visually consistent with the actual image. 
Thus, features matching between the real and rendered images are found at the same positions (up to noise), as can be see from the (short) green lines. %
\Fig\ref{fig:project_old_ref_poses} shows selected examples where the original reference poses were rather inaccurate. 
Visual comparison between the original and our refined poses showed that our approach consistently produced more accurate poses for all nighttime queries.

It is also worth noting that D2-Net features can provide robust matches even though the rendered images (using a model reconstructed from daytime imagery) are visually very different from the actual images and contain non-trivial rendering artifacts.

\PAR{Quantitative evaluation.} 
To quantify the differences between the original and our reference poses, we computed the differences in camera position and orientation (see \eqref{eq:pose_diff}). 
\Fig\ref{fig:refined_gt_pose_change} shows the results of this comparison.
It can be seen that there exists a non-trivial discrepancy between the original and refined reference poses. %

\cite{Sattler2018cvpr} measures localization accuracy by the percentage of nighttime query poses estimated within (0.5 \si{m}, 2 \si{deg}), (1 \si{m}, 5 \si{deg}), and (5 \si{m}, 10 \si{deg}) of the reference poses. 
These thresholds are also shown in \Fig\ref{fig:refined_gt_pose_change}. 
As can be seen, the differences between the original and refined poses fall outside of the largest error threshold for 11 images ($\sim 11.2\%$ of all the nighttime queries). %
Interestingly, the best results reported on \href{https://www.visuallocalization.net/}{visuallocalization.net} register 88.8\% of the nighttime queries within 5 \si{m} and 10 \si{deg}. 
Thus, state-of-the-art methods might actually be more accurate than the reference poses.

Finally, \Tab\ref{tab:eval_original_images} evaluates several state-of-the-art localization methods using the existing and refined reference poses. %
As can be seen, the accuracy of the localization methods is indeed (significantly) under-estimated by the existing reference poses of the nighttime images in the Aachen Day-Night dataset. %
In contrast, our reference poses allow us to measure localization performance more accurately. 
\Tab\ref{tab:eval_original_images} also provides results for additional evaluation measures for our new reference poses, which will be discussed in \Sec\ref{subsec:exp_stoa_diff_metrics}.
Note that the improvement reported here is particular to the nighttime images in the Aachen Day-Night dataset. The result on a different dataset will depend on the quality of the existing reference poses in the dataset.

\PAR{Summary.} 
Our results clearly show that our new reference poses are more accurate than the original poses. 
We will integrate our new poses in the \href{https://www.visuallocalization.net/}{visuallocalization.net} online benchmark, allowing us to easily update all results on the website.

\begin{figure*}[th!]
    \centering
    \includegraphics[width=0.95\linewidth]{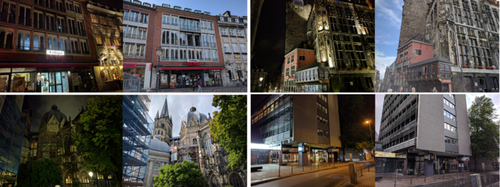}
    \caption{Pairs of day-night images taken from similar poses. 
    We obtain reference poses for the daytime images via SfM. 
    The resulting poses are used to initialize our approach for generating poses for the nighttime images.
    For the SIFT and D2-Net registration baselines, an additional $20$ daytime images that overlap with these images are selected from the base model for the daytime image in each pair. 
    }
    \label{fig:day_night_matches_extension}
\end{figure*}

\subsection{Extending the Aachen Day-Night Dataset}
\label{subsec:exp_extending_aachen}

Our approach is capable of estimating an accurate pose from a coarse initialization. 
Besides verifying and refining existing reference poses, our approach can also be used for generating reference poses for new images. %
In the next experiment, we thus extend the Aachen Day-Night dataset by additional nighttime queries. 
We compare our reference poses with two registration baselines using SIFT and D2-Net features, respectively.

\PAR{Reference pose generation.} 
As shown in \Fig\ref{fig:day_night_matches_extension}, 
we captured a daytime photo from a similar pose for each the 119 new nighttime images. 
The poses of these daytime images in the extended base model, obtained via SfM, then provide initial pose estimates for the nighttime queries that are subsequently refined by our approach. %
We excluded images for which our method resulted in less than $10$ effective inliers to avoid unreliable reference poses. 
This results in reference poses for $93$ out of the $119$ images.

We compare our method with two baselines using SIFT and D2-Net features, respectively.
Both baselines match features between the 93 new nighttime queries and a small set of images in the extended base SfM  model. 
For a nighttime query, this set includes the corresponding daytime image $\mathcal{I}_D$ as well as the $20$ images in the extended base model that share the most 3D points with $\mathcal{I}_D$. 
2D-2D matches between the nighttime image and the daytime photos in the set then yield a set of 2D-3D correspondences based on the 3D points visible in the latter. 
COLMAP's image registration pipeline was then used to obtain the camera pose based on these matches. 
Note that for D2-Net features, we re-triangulated the extended base 3D model before day-night feature matching.

\begin{figure}[th!]
    \centering
    \begin{subfigure}{0.99\linewidth}
    \centering
    \includegraphics[width=\linewidth]{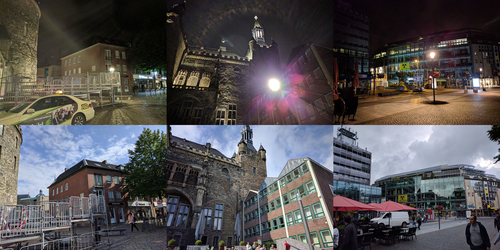}
    \caption{Night-day matches for which SIFT registration failed.}
    \end{subfigure}
    \begin{subfigure}{0.99\linewidth}
    \centering
    \includegraphics[width=\linewidth]{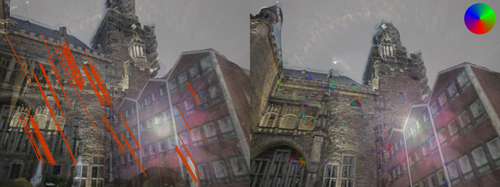}
    \includegraphics[width=\linewidth]{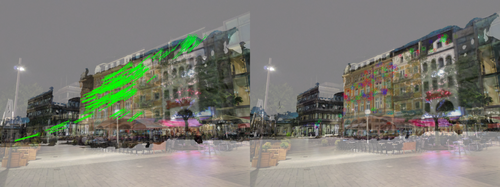}
    \caption{Refinement process of our method (1st iteration).}
    \label{fig:ours_on_sift_fail}
    \end{subfigure}
    \caption{Typical failure cases of the SIFT registration baseline.
    \textbf{Top}: nighttime images where SIFT registration failed and the corresponding daytime images;
    \textbf{Bottom}: Visualization of the first iteration of our method (left: initial pose; right: refined pose). The differences between D2-Net features and the projection of the matching 3D points are color coded according to the direction in the image plane (cf. legend in the top-right).
    }
    \label{fig:extension_sift_fail}
\end{figure}

\PAR{Robustness.}
Both the D2-Net baseline and our method are able to consistently estimate poses for challenging images for which the SIFT baseline fails. %
\Fig\ref{fig:extension_sift_fail} shows such failure cases of SIFT. 
In each of the shown cases, there is a strong light source in the scene, causing significant appearance differences between the day and nighttime images. 
SIFT is not able to deal with these strong changes. %
In contrast, our method, as well as the D2-Net baseline, which relies on high level learned features, are able to handle these cases (cf. \Fig\ref{fig:ours_on_sift_fail}).

\cite{Sattler2018cvpr} reported that the reference poses obtained via SfM and SIFT were unreliable. 
Interestingly, we observe the opposite for many images in our experiments. 
We attribute this to the inclusion of the corresponding daytime images: 
as shown in~\cite{Torii2018pami}, SIFT features better handle day-night changes under small viewpoint changes. %
Note that daytime images taken from very similar poses are not available for the original nighttime queries.

\begin{figure}[t]
    \centering
    \includegraphics[width=\linewidth]{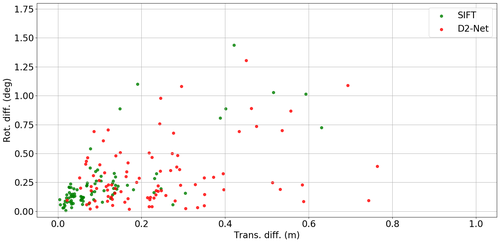}
    \caption{Distribution of the pose difference between our method and the two registration baselines.}
    \label{fig:extension_pose_diff_overall}
\end{figure}

\PAR{Quantitative evaluation.}
Excluding the failure cases, we computed the pose differences between our method and two baselines.
The results of this comparison are shown in \Fig\ref{fig:extension_pose_diff_overall}.
Interestingly, the poses from our method and the SIFT registration are very consistent.
For the majority of the images, the pose difference is below $0.2$ \si{m} and $0.5$ \si{deg}.
In contrast, we observe much larger difference between our poses and the D2-Net registration baseline.
As there is no external reference poses that can be used to calculate the absolute pose accuracy, we resort to visual inspection based on the renderings. %

\begin{figure}[t]
    \centering
    \includegraphics[width=\linewidth]{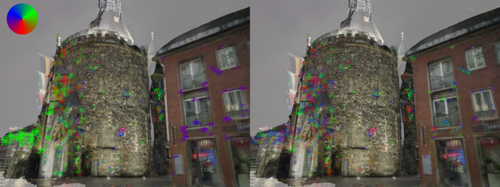}
    \includegraphics[width=\linewidth]{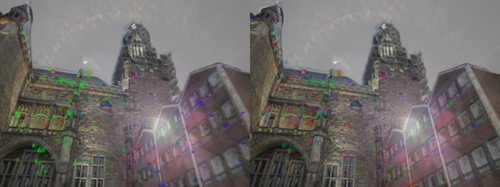}
    \includegraphics[width=\linewidth]{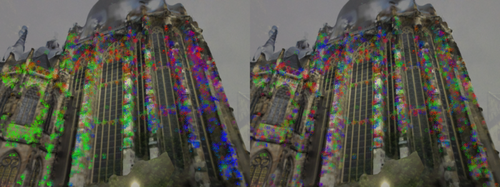}
    \caption{
    Comparing the D2-Net baseline against our refinement. %
    \textbf{Left}: overlay of real photos and images rendered with D2-Net poses. D2-Net features in the rendered images are connected to the matching locations in the real images (circles), and the color indicates the direction of the feature location differences in the two images (see legend in the top-left).
    \textbf{Right}: corresponding visualization using poses obtained by one iteration of our method (initialized with D2-Net poses).
    The patterns of the feature directions in the left images indicate the inaccuracy in the poses from D2-Net registration, which are corrected with our method (right images).
    }
    \label{fig:d2net_fail}
\end{figure}

\PAR{Visual inspection.} 
\Fig\ref{fig:d2net_fail} analyses example poses obtained by the D2-Net baseline. 
Besides overlaying the real and rendered images, we also show D2-Net features matches between the two. %
For each match, we compute the 2D offset between the feature positions in the real and the rendered view. 
Following~\cite{Schoeps2019CVPR}, we color-code the features based on the directions of these 2D offsets. %
As argued in~\cite{Schoeps2019CVPR}, these directions should be randomly distributed for accurate pose estimates. 
Patterns of similar direction in the same region of an image indicate a shift between the two images and thus pose errors.

The D2-Net poses in \Fig\ref{fig:d2net_fail} are visually more accurate than those in \Fig\ref{fig:project_old_ref_poses} and \Fig\ref{fig:ours_on_sift_fail}. 
Still, we observe clear patterns in the distribution  of the directions (\eg the concentration of green color on one side and purple on the other), which indicates  inaccuracies in the poses of the D2-Net baseline.
We further used one iteration of our method to refine the D2-Net poses. 
As can be seen in \Fig\ref{fig:d2net_fail}, the refinement improves the distribution of directions. 
We conclude that our approach is able to provide more accurate than the D2-Net baseline.

\begin{figure}
    \centering
    \begin{subfigure}{0.99\linewidth}
    \centering
    \includegraphics[width=0.32\linewidth]{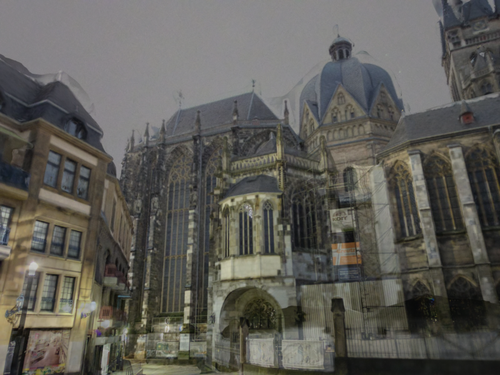}
    \includegraphics[width=0.32\linewidth]{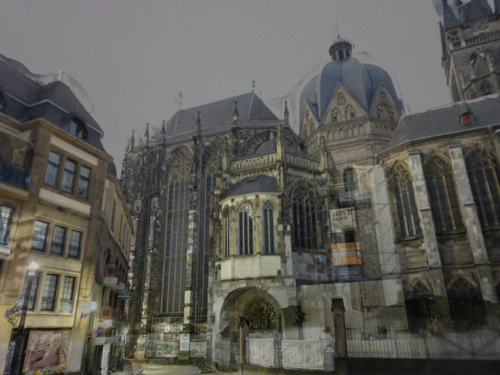}
    \includegraphics[width=0.32\linewidth]{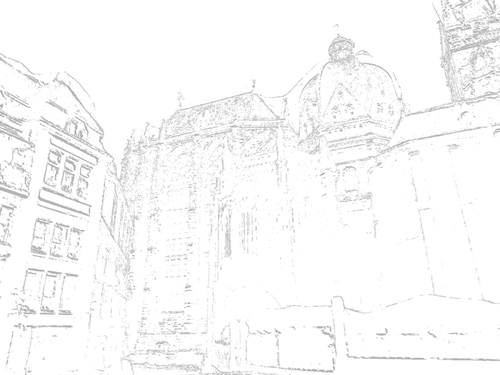}
    \caption{Typical visual difference between poses the SIFT baseline and our method.
    Left and Middle: overlay of the rendered and real images for SIFT respectively our poses.
    Right: the intensity difference of the rendered images. The images are converted to 8-bit gray-scale images, and the pixels with intensity difference larger than $10$ are shown in gray.}
    \label{fig:typical_visual_sift_ours}
    \end{subfigure}
    \begin{subfigure}{0.99\linewidth}
    \centering
    \includegraphics[width=0.49\linewidth]{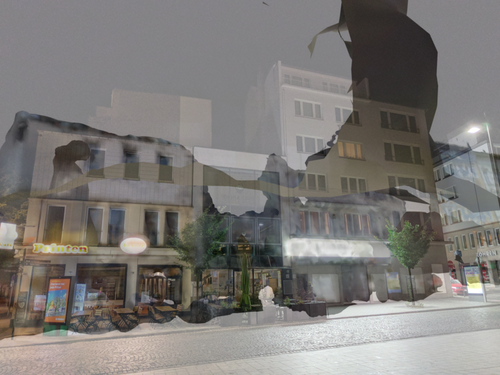}
    \includegraphics[width=0.49\linewidth]{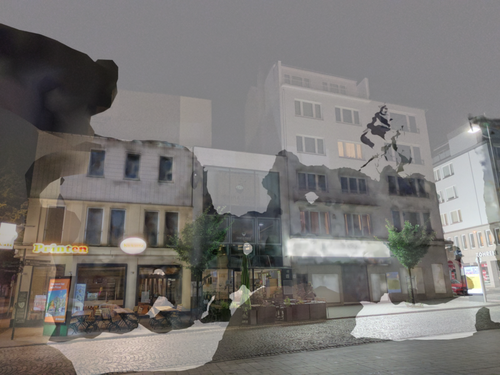}
    \includegraphics[width=0.49\linewidth]{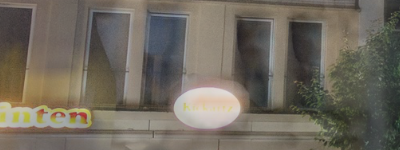}
    \includegraphics[width=0.49\linewidth]{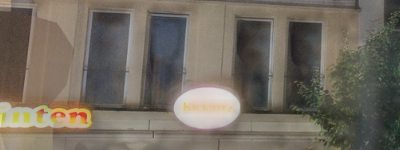}
    \caption{Example where the pose from our method (right) is more accurate than the SIFT pose  (left), as can be seen from the sign in the middle of the cutouts.}
    \label{fig:visual_diff_better}
    \end{subfigure}
    \begin{subfigure}{0.99\linewidth}
    \centering
    \includegraphics[width=0.49\linewidth]{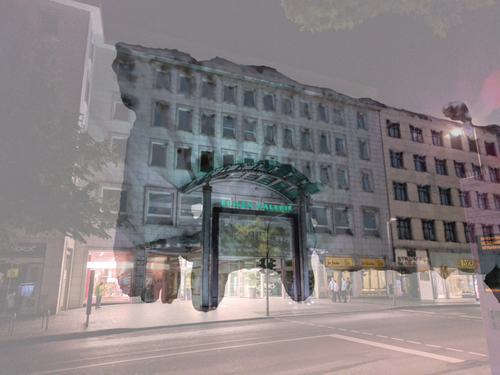}
    \includegraphics[width=0.49\linewidth]{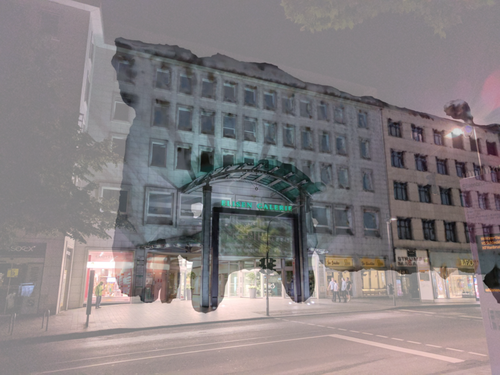}
    \includegraphics[width=0.49\linewidth]{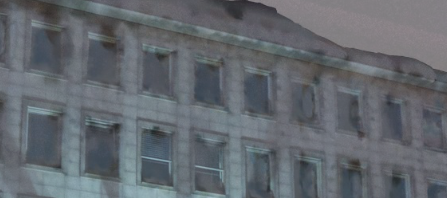}
    \includegraphics[width=0.49\linewidth]{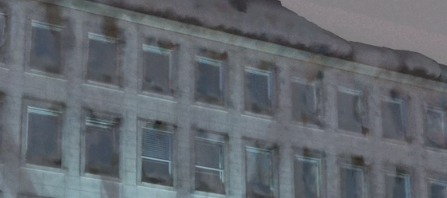}
    \caption{Example where the SIFT pose (left) is more accurate than our method (right), as can be seen from the windows and the edge of the roof in the cutouts.
    }
    \label{fig:visual_diff_worse}
    \end{subfigure}
    \caption{Visual comparison of images rendered from the poses obtained by our method and the SIFT baseline.}
    \label{fig:extension_sift_vs_ours_eg}
\end{figure}

As can be seen from \Fig\ref{fig:extension_pose_diff_overall}, the pose differences between our approach and the SIFT baseline are significantly smaller than the differences between our approach and D2-Net. 
Unlike for D2-Net poses, we did not see strong feature direction patterns for the SIFT poses. 
We therefore omit the corresponding visualizations. %
We observe that if the SIFT baseline is able to estimate a pose it is usually visually similar to the pose obtained with our approach (cf. \Fig\ref{fig:typical_visual_sift_ours}).
There are images where the poses from our method seem to be visually more accurate than the SIFT registration and vice versa (shown in \Fig\ref{fig:visual_diff_better} and \ref{fig:visual_diff_worse}, respectively).
Yet, overall there are only 7 out of the 93 new nighttime queries for which we consider the SIFT poses to be visually more accurate than the poses provided by our method. 
For these images, we use the SIFT poses as reference poses. 
At the same time, SIFT failed to provide poses for 5 of the nighttime images due to a lack of sufficient matches.

\PAR{Discussion and summary.} 
It is interesting to see that SIFT poses are not necessarily more accurate than our poses. 
SIFT features are much more accurately localized in images than D2-Net features~\cite{Dusmanu19cvpr}. 
Thus, one might have expected that a few accurately localized SIFT matches are better than many less accurately localized D2-Net matches. 
Yet, finding more matches with D2-Net between the renderings and the real images seems to compensate for the inaccuracy of the D2-Net feature detections.

For the newly acquired nighttime images, we observe that our approach performs similar to SIFT in terms of accuracy. 
In this case, SIFT benefits from daytime images taken from similar viewpoints. 
As evident from the failure cases of SIFT on both the original and new queries, our approach is more robust than the SIFT baseline. 
As a result, our approach is better suited to for reference pose generation for datasets that benchmark long-term visual localization algorithms.

Compared with the D2-Net baseline, the poses resulting from our method are more accurate. %
The main difference between the D2-Net baseline and our approach is the use of rendered images. 
The results thus validate our choice to iteratively render the scene from the current pose estimate and match features against the rendering. 
Moreover, as seen from the analysis of the D2-Net baseline, the ability of our method to verify and refine existing poses is also valuable when it is combined with other approaches.

\subsection{Ablation Study}
\label{subsec:exp_ablation}
Next, we present ablation studies to analyze our proposed approach. 
We first obtain an estimate for the stability of our reference poses. 
Next, we determine the impact of using different features and rendering techniques, which are the two key ingredients in our method.
Finally, we show failure cases of our method.

\PAR{Pose stability.}
To provide a quantitative measure of the uncertainties/stability of the reference poses obtained with our method, we compute the {sampling uncertainties} as described in \Sec\ref{subsec:uncertainty_quantification} for both the original and additional nighttime images: 
we randomly sample a percentage of 2D-3D matches from the inliers used to estimate the reference poses. 
This sample is then used to obtain another pose estimate. %
The differences between these new and our poses provide a measure for the stability of the minima found by our approach. 

We used three sampling rates that use 90\%, 50\%, and 10\% of the inliers, respectively. 
For each rate, we drew 50 random samples and report the median position and orientation differences. 
In addition, since our method uses different rendering techniques and is an iterative process, we also computed the following for comparison:
\begin{itemize}
    \item \emph{Compare-InvSfM}: the differences between the refined poses using both types of rendered images and using InvSfM only;
    \item \emph{Compare-Mesh}: the differences between the refined poses using both types of rendered images and mesh rendering only;
    \item \emph{Compare-Prev-Iter}: the pose differences between the two last iterations of our refinement process.
\end{itemize}

\begin{figure}[t]
    \centering
        \includegraphics[width=\linewidth]{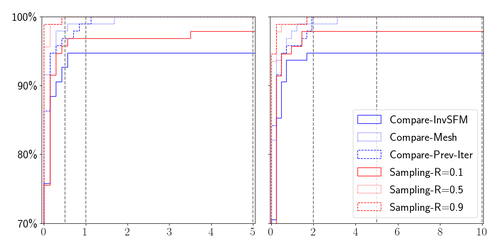}
        \includegraphics[width=\linewidth]{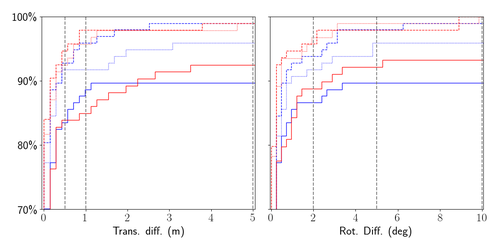}
    \caption{Different uncertainties for the original (\textbf{top}) and additional (\textbf{bottom}) Aachen Night images.
    The vertical dash lines corresponds to the error thresholds proposed in~\cite{Sattler2018cvpr} and used by the online benchmark.}
    \label{fig:pose_uncertainties}
\end{figure}

The results of our comparisons are shown in \Fig\ref{fig:pose_uncertainties}.
For the original images, more than $90\%$ of the images are below the finest error threshold (0.5 \si{m}, 2 \si{deg}) of the visual localization benchmark, independently of which sampling rate and rendering is used. 
For the additional images, the uncertainties are higher. 
Still, more than $80\%$ of the images fall in that threshold as well.
The fact that the uncertainties of the additional images are overall higher than the original images indicates that the newly added images might be more challenging. %
Regarding the different rendering techniques, images rendered using the MVS mesh seem to provide more information for the final refined poses, as \emph{Compare-Mesh} shows less uncertainty than \emph{Compare-InvSfM}.

While it is difficult to quantify the absolute uncertainties, the uncertainties shown in \Fig\ref{fig:pose_uncertainties} indicate that the reference poses generated using our method are at least stable solutions considering the available 2D-3D matches. 
This can be seen from the fact that even using as little as 10\% of the available inlier matches leads to very similar pose estimates for nearly all images.

\begin{figure*}[t]
    \centering
    \begin{subfigure}{0.77\textwidth}
    \centering
    \includegraphics[width=\linewidth]{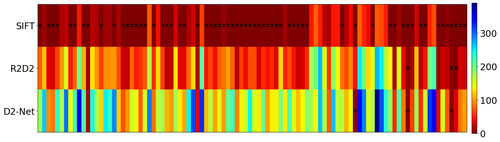}
    \end{subfigure}
    \begin{subfigure}{0.22\textwidth}
    \centering
    \includegraphics[width=\linewidth]{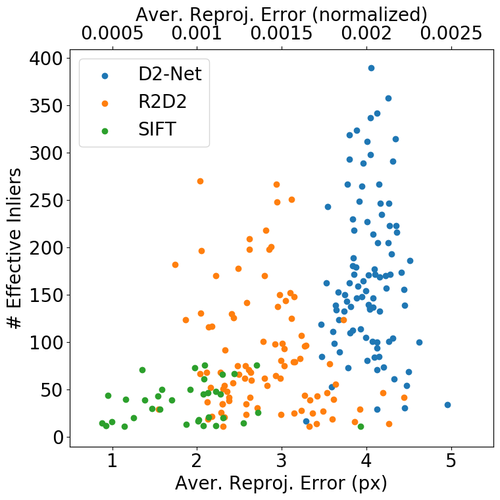}
    \end{subfigure}
    
    \begin{subfigure}{0.77\textwidth}
    \centering
    \includegraphics[width=\linewidth]{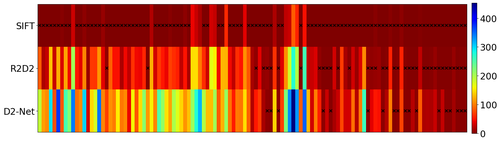}
    \end{subfigure}
    \begin{subfigure}{0.22\textwidth}
    \centering
    \includegraphics[width=\linewidth]{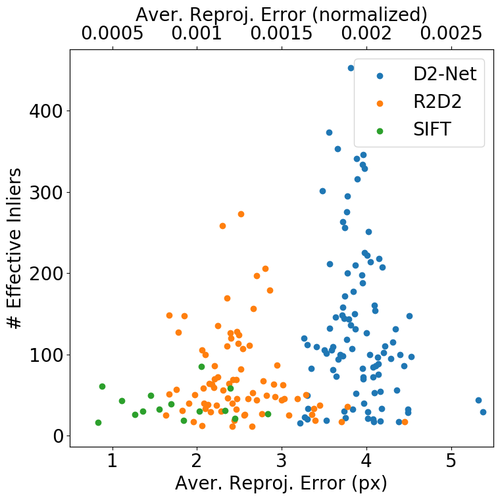}
    \end{subfigure}
    \caption{Effect of using different features in our method.
    \textbf{Left}: the number of effective inliers for each image. Each block along the horizontal axis corresponds to one image. A black cross indicates there are less than 10 effective inliers, \ie the pose is likely not reliable.
    \textbf{Right}: the number of effective inliers and the mean reprojection error (after nonlinear optimization) for different features. Failure cases (\ie the black crosses) are excluded.
    The normalized reprojection error is normalized by the image diagonal length.
    The top row shows the result for the original Aachen nighttime images, and the bottom for  additional images.
    }
    \label{fig:ablation_ftr_compare}
\end{figure*}

\PAR{Different features.} 
Instead of using D2-Net features, we also used 
SIFT and R2D2~\cite{Revaud19neurips} features to obtain matches between the rendered and real images.

\Fig\ref{fig:ablation_ftr_compare} compares the results obtained with different types of features.
As can be seen, SIFT failed to find enough matches in most cases for both the original and additional night images. 
This is not surprising considering SIFT relies on low-level image statistics, which are strongly impacted by imperfections in the MVS model and the invSfM rendering process. %
In contrast, both D2-Net and R2D2 features were able to find enough matches for most of the original Aachen night images.
The success rate for both features drops on the additional Aachen night images, where the D2-Net feature performed better.
Plotting the reprojection error (after nonlinear optimization) against the number of effective inliers, we observe a clear trend across different features: 
D2-Net recovers the most matches, followed by R2D2 and SIFT; while SIFT features were most accurately localized in the images, D2-Net has the largest reprojection errors.

\begin{figure}[t!]
    \centering
        \includegraphics[width=\linewidth]{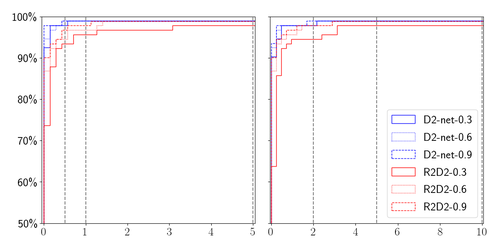}
        \includegraphics[width=\linewidth]{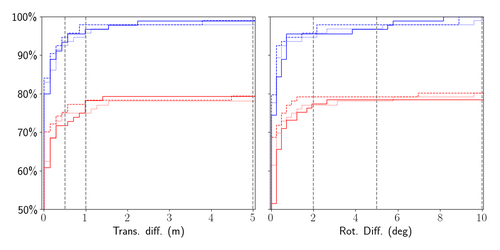}
    \caption{Sampling uncertainties of the D2-Net and R2D2 poses for the original (\textbf{top}) and additional (\textbf{bottom}) Aachen night images. Median position and orientation errors over 50 random samples are shown.}
    \label{fig:ablation_d2net_r2d2_uncer}
\end{figure}

To see how the number of effective inliers and reprojection error translates to the quality of the refined poses, we further computed the {sampling uncertainties} for D2-Net and R2D2, shown in \Fig\ref{fig:ablation_d2net_r2d2_uncer}. We excluded SIFT since it failed for most of the images.
It can be seen that the refined poses from D2-Net features are more stable than the R2D2 poses for both the original and additional images. 

The results validate our choice of using D2-Net features to match between real and rendered images as they better handle imperfections in the renderings. 

\begin{figure}[t]
    \centering
    \begin{subfigure}{0.99\linewidth}
    \centering
    \includegraphics[width=\linewidth]{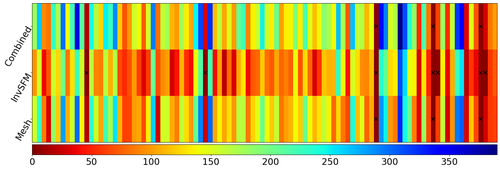}
    \caption{Original Aachen night images.}
    \end{subfigure}
    \begin{subfigure}{0.99\linewidth}
    \centering
    \includegraphics[width=\linewidth]{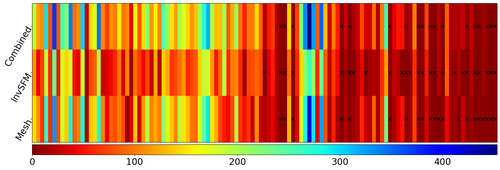}
    \caption{Additional Aachen night images.}
    \label{fig:ablation_mesh_nn_additional_aachen}
    \end{subfigure}
    \caption{The number of effective inliers for D2-Net features when different rendering techniques are used.
    The visualization is the same as \Fig\ref{fig:ablation_ftr_compare}.}
    \label{fig:ablation_nn_mesh}
\end{figure}

\PAR{Different rendering techniques.} 
The experiments presented so far used both rendering types (using MVS mesh and the invSfM process). 
Next, we compare using both types against using only one of the two using the number of effective inliers. 

As can be seen in \Fig\ref{fig:ablation_nn_mesh}, using renderings based on the MVS mesh in general resulted in more effective inliers compared to using  invSfM for rendering. 
Accordingly, there are more images where our method could find sufficient effective inliers in the images rendered from mesh.
This is also consistent with our results in \Fig\ref{fig:pose_uncertainties}, which show that the poses based only on mesh rendering are more accurate than those obtained using only invSfM. 
Yet, there are a few cases where mesh rendering fails while invSfm rendering succeeds. 
The corresponding nighttime images show parts of the model that are only sparsely covered by images and where the MVS reconstruction is thus incomplete. 
The invSfM process seems to be more stable for such cases.

Combining the 2D-3D matches obtained from both types of renderings increases the number of effective inliers. 
Note that the effective inlier count selects at most one inlier for each 50 pixels by 50 pixels region in an image. 
A higher effective inlier count thus indicates that the matches found by the two rendering types are somewhat complimentary as matches are found in different image regions. 
Moreover, there are a few cases (right part of \Fig\ref{fig:ablation_mesh_nn_additional_aachen}) for which using both rendering types is necessary to obtain sufficiently many inliers. 

The results validate our choice of using both rendering techniques as they are (partially) complimentary.

\begin{figure}[t]
    \centering
    \begin{subfigure}{0.49\linewidth}
    \centering
    \includegraphics[width=\linewidth]{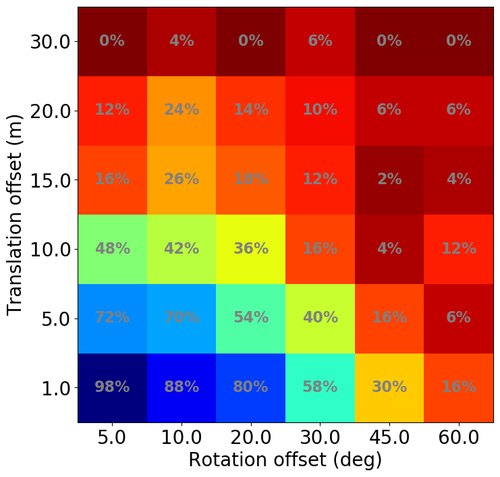}
    \caption{After the 1st iteration}
    \label{fig:abl_sensitivity_first}
    \end{subfigure}
    \begin{subfigure}{0.49\linewidth}
    \centering
    \includegraphics[width=\linewidth]{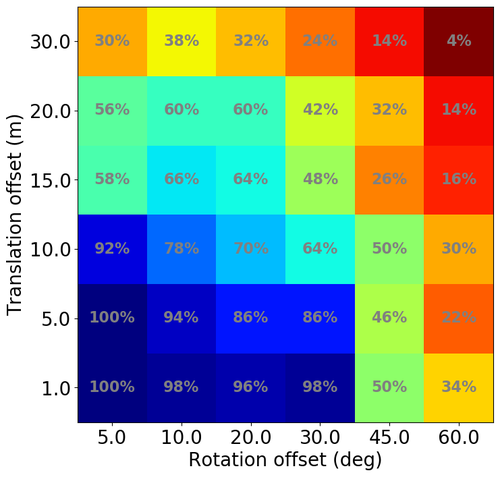}
    \caption{After the 5th iteration}
    \label{fig:abl_sensitivity_last}
    \end{subfigure}
    \caption{
    Success rates of our method in the presence of different initial error levels on the original Aachen nighttime images using D2-Net features.
    Each cell corresponds to the success rate of 50 trials of pose refinement using perturbed poses as input.
    A perturbed pose was generated by adding both random rotation and position offsets of certain magnitudes, as indicated on the $x$ and $y$ axes respectively, to a refined pose.
    The refinement of a pose is considered successful if the pose error after refinement is within $0.25$ m and $1.0$ deg.
    }
    \label{fig:ablation_sensitivity}
\end{figure}
\PAR{Sensitivity to initialization.} 
Our approach requires an initial pose estimate as input to the iterative refinement. 
Naturally, our approach will fail if the initial pose estimate is not accurate enough. 
To determine the sensitivity of our approach to the initial pose error, we randomly perturbed the refined poses by translations and rotations of different magnitudes and used the perturbed poses as input to our method.
We then measured the sensitivity of our method to the initial pose error by the success rates at different perturbation levels, and the refinement is considered successful if the manually added error can be reduced to under $0.25$ m and $1.0$ deg, which is half of the smallest fixed error threshold used in our evaluation (cf. \Sec\ref{subsec:exp_stoa_diff_metrics}).
The experiment was performed on the original Aachen nighttime images using D2-Net features.
As can be seen in \Fig\ref{fig:abl_sensitivity_last}, 
the success rate of our method is more than $50\%$ for initial pose errors up to $30$ deg and $10$ m.
The success rate is more than $90\%$ for disturbances within $10$ deg and $5$ m.
Moreover, the increase of the success rates from the 1st iteration (\Fig\ref{fig:abl_sensitivity_first}) to the $5$th iteration (\Fig\ref{fig:abl_sensitivity_last}) indicates the necessity of performing the refinement iteratively multiple times.
Therefore, our method is quite robust to the errors in initial poses and can be potentially used with systems that provide less accurate localization information (\eg GPS) to get more accurate poses automatically.

\begin{figure}[t]
    \centering
    \begin{subfigure}{0.99\linewidth}
    \centering
    \includegraphics[width=0.49\linewidth]{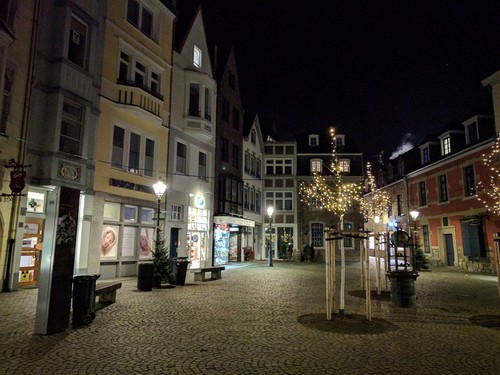}
    \includegraphics[width=0.49\linewidth]{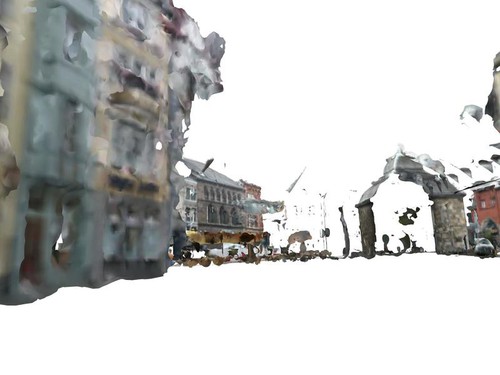}
    \caption{Failure mode 1: low quality mesh.}
    \label{fig:fail_mesh}
    \end{subfigure}
    \begin{subfigure}{0.99\linewidth}
    \centering
    \includegraphics[width=0.49\linewidth]{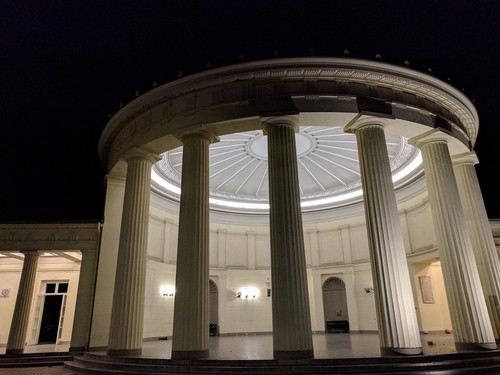}
    \includegraphics[width=0.49\linewidth]{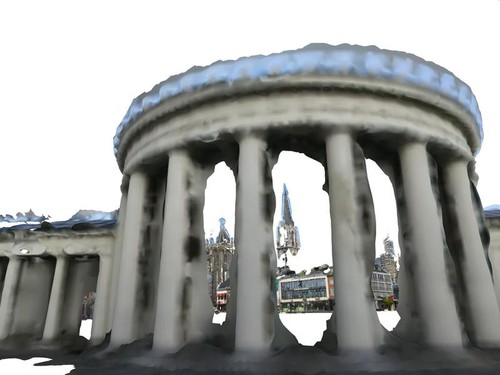}
    \caption{Failure mode 2: scene with little texture.}
    \label{fig:fail_textureless}
    \end{subfigure}
    \caption{Typical failure cases of our method. Left: real nighttime images; Right: MVS mesh renderings from the initial pose.}
    \label{fig:ablation_failures}
\end{figure}

\PAR{Failure cases.}
\Fig\ref{fig:ablation_failures} shows examples of two typical failure cases of our method. %
The first failure mode is when the nighttime image was taken in a part of the scene where the MVS mesh is of low quality, \eg parts of the surface have not been reconstructed (cf. \Fig\ref{fig:fail_mesh}).
This could be overcome by using a more complete/higher quality mesh of the scene, but might require additional data capture.
The second failure mode is caused by weakly textured scenes %
(cf. \Fig\ref{fig:fail_textureless}).
In the shown example, the rendered image %
is of reasonable quality visually.
However, due to the lack of texture, our method failed to find enough matches between the rendered image and the real night image.
Using contour edges as an additional feature type could help avoid this failure mode. 
However, edges are also typically harder to match than local features. 
Furthermore, care would need to be taken to handle protruding regions in the MVS model.

\begin{table*}[t]
    \caption{Localization accuracy using different metrics on the extended Aachen Day-Night dataset. We compare the methods based on the Pose Error, \ie the percentage of queries localized within fixed error thresholds of the reference poses. As can be seen, our more accurate reference poses yield a better measure of pose accuracy. For our poses, we also report results for two additional metric: the percentage of queries localized within sampling-based thresholds (Sampling) of the reference poses (cf. \Sec\ref{sec:loc_metrics:direct}) and the percentage of queries with maximum reprojection errors within given error thresholds in pixels  (Reprojection Diff.) (cf. \Sec\ref{sec:loc_metrics:indirect}).}
    \label{tab:eval_overall}
    \centering
    \setlength{\tabcolsep}{1pt}
    \footnotesize{
    \begin{tabular}{l | c| c| c| c| c |c }
        \toprule
         &  \multicolumn{3}{c|}{\textbf{Original Night Images}} &
         \multicolumn{3}{c}{\textbf{All Night Images}}\\
         & \multicolumn{1}{c}{\begin{tabular}{@{}c@{}}Pose Error\\ \scriptsize{0.25\si{m},2$^\circ$/0.5\si{m},5$^\circ$/5\si{m},10$^\circ$}\end{tabular}} 
         & \multicolumn{1}{c}{\begin{tabular}{@{}c@{}}Sampling \\ \scriptsize{(50\%/30\%/10\%)} \end{tabular}}  
         & \begin{tabular}{@{}c@{}}Reprojection Diff. \\ \scriptsize{(10/20/50/100 px)} \end{tabular}
         & \multicolumn{1}{c}{\begin{tabular}{@{}c@{}}Pose Error\\ \scriptsize{0.25\si{m},2$^\circ$/0.5\si{m},5$^\circ$/5\si{m},10$^\circ$}\end{tabular}} 
         & \multicolumn{1}{c}{\begin{tabular}{@{}c@{}}Sampling \\ \scriptsize{(50\%/30\%/10\%)} \end{tabular}}
         & \begin{tabular}{@{}c@{}}Reprojection Diff. \\ \scriptsize{(10/20/50/100 px)} \end{tabular} \\
         \midrule
         D2-Net 
         & 84.7/90.8/98.0
         & 11.2/19.4/43.9
         & 56.1/80.6/92.9/95.9
         & 71.7/90.6/97.9
         & 6.3/11.0/30.9
         & 36.1/73.8/91.1/96.9\\
         R2D2-20k
         & 82.7/90.8/95.9
         & 7.1/11.2/38.8
         & 54.1/76.5/89.8/93.9
         & 73.3/88.5/96.3
         & 5.2/7.9/29.8
         & 40.8/72.8/91.6/94.8\\
         R2D2-40k
         & 80.6/91.8/98.0
         & 7.1/13.3/44.9
         & 56.1/76.5/92.9/95.9
         & 73.3/88.5/97.9
         & 5.8/8.9/33.0
         & 41.9/73.3/91.6/95.8\\
         \bottomrule
    \end{tabular}
    }%
\end{table*}

\subsection{Evaluation of State-of-the-Art Methods}
\label{subsec:exp_stoa_diff_metrics}
\Tab\ref{tab:eval_original_images} evaluates published state-of-the-art localization methods using our new reference poses for the original nighttime images. 
The results were obtained by re-evaluating poses submitted to \href{https://www.visuallocalization.net/}{visuallocalization.net}.\footnote{There results available at \href{https://www.visuallocalization.net/}{visuallocalization.net} for methods that outperform the approaches used in \Tab\ref{tab:eval_original_images}. For our experiments, we limited ourselves to methods that have been published in peer reviewed conferences and journals. Updated results for the other methods will be available on the benchmark website once we update the reference poses.} 
In the following, we present results for state-of-the-art methods on our new extended Aachen Day-Night dataset. 
Note that the extended dataset uses a larger reference SfM model than the original one and we thus cannot use results from the benchmark website. 

Given that D2-Net and R2D2 features achieve state-of-the-art results in \Tab\ref{tab:eval_original_images}, we use two image retrieval-based approaches based on these features in our evaluation. 
Both approaches first re-triangulate the reference SfM model with feature matches between the reference images found by D2-Net respectively R2D2. 
Next, NetVLAD~\cite{Arandjelovic2016CVPR} is used to retrieve the 20 most similar reference image for each nighttime query. 
Feature matches between each query and its retrieved image yield a set of 2D-3D matches via the 3D points visible in the reference images. 
These 2D-3D matches are used for pose estimation against the reference model inside COLMAP.\footnote{Based on code available at \url{https://github.com/tsattler/visuallocalizationbenchmark}.} 
For R2D2, we provide results for two variants that use at most 20k (R2D2-20k) respectively 40k (R2D2-40k) features per image.

\Tab\ref{tab:eval_overall} shows the results of our experiments using the evaluation measures discussed in \Sec\ref{sec:loc_metrics}. 
Similarly, \Tab\ref{tab:eval_original_images} also shows results for all metrics for our new reference poses. 
Overall, %
the accuracy is lower when considering all nighttime queries compared to only focusing on the original night images, independent of the metric used. %
This indicates the newly added images might be more challenging. 
In the following, we discuss the results per evaluation metric.

\PAR{Pose error with fixed thresholds.} 
We first considered the three fixed error thresholds used in~\cite{Sattler2018cvpr} and on the benchmark website, \ie (0.5 \si{m}, 2 \si{deg}), (1 \si{m}, 5 \si{deg}), and (5 \si{m}, 10 \si{deg}). 
Based on this metric, %
the performance on the original and extended Aachen dataset seems saturated for certain algorithms (\eg D2-Net and R2D2).
However, these thresholds were originally chosen to take the uncertainties in the original nighttime reference poses into account. %
As shown in our previous experiments, our new reference poses are significantly more accurate. 
As such, using rather loose thresholds could lead to an overestimate in the localization accuracy,
and we instead chose a set of tighter thresholds \ie (0.25 \si{m}, 2 \si{deg}), (0.5 \si{m}, 5 \si{deg}), and (5 \si{m}, 10 \si{deg}) for evaluation.
Furthermore, as discussed in \Sec\ref{sec:loc_metrics}, using the same thresholds for all images does not take into account that the uncertainty in the pose depends on the distance of the camera to the scene.

\begin{figure}
    \centering
    \begin{subfigure}{\linewidth}
    \centering
    \includegraphics[width=\linewidth]{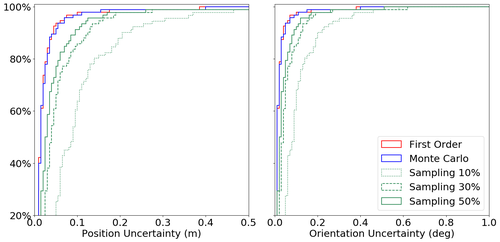}
    \caption{
    Cumulative histograms of the position and rotation uncertainties computed using different methods. Curves on top indicate lower uncertainty estimates.
    }
    \label{fig:diff_uncer_dist}
    \end{subfigure}
    \begin{subfigure}{\linewidth}
    \centering
    \includegraphics[width=\linewidth]{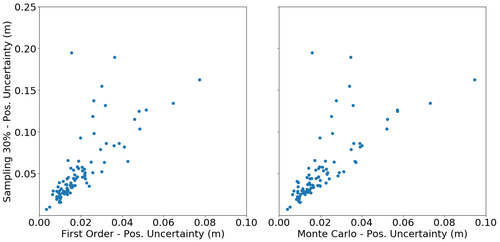}
    \caption{
    Correlation between different position uncertainties.
    \textbf{Left}: $30\%$ Sampling uncertainty with respect to the first order uncertainty.
    \textbf{Right}: $30\%$ Sampling uncertainty with respect to the Monte Carlo uncertainty.
    }
    \label{fig:diff_uncer_cor}
    \end{subfigure}
    \caption{
    Comparison of the different uncertainty definitions provided in \Sec\ref{subsec:uncertainty_quantification} on the original Aachen nighttime images using the refined poses.
    }
    \label{fig:diff_uncertainties}
\end{figure}

\PAR{Per image error thresholds.}
The second metric aims at computing error thresholds on the camera pose per image. 
We first show the results using the sampling uncertainties \eqref{eq:sampling_uncertainties} as error thresholds and then discuss the first order and Monte Carlo uncertainties \eqref{eq:uncertainty_first_order} and \eqref{eq:mc_uncertainties}.
For each reference pose, we randomly sampled %
set containing $10\%$, $30\%$ and $50\%$ of the inliers of our method. 
For each sampling percentage, we drew 50 samples and computed the median position and orientation difference between the poses obtained from the samples and the reference poses. 
These median differences were then used as the error thresholds. 
As can be seen from \Tab\ref{tab:eval_original_images} and \Tab\ref{tab:eval_overall}, 
the sampling uncertainties tend to under-estimate the localization performance of the different methods. 
This is due to the fact that our reference poses are rather stable under using a subset of the inlier matches (cf. \Fig\ref{fig:ablation_d2net_r2d2_uncer}). 
The sampling uncertainties reflect the stability of the local minimum reached in the refinement process, rather than the absolute uncertainties. 
Thus, this metric should not be used to evaluate localization performance.

As for the first order and Monte Carlo uncertainties, we found that they tend to be lower than the sampling uncertainties.
As an example, a comparison of different uncertainties on the original Aachen nighttime images (using the refined pose) is shown in \Fig\ref{fig:diff_uncertainties}.
From \Fig\ref{fig:diff_uncer_dist}, we can see that the uncertainties estimated by first order and Monte Carlo methods are in general lower than the sampling uncertainties, even for the highest sampling ratio of $50\%$.
We also inspected the correlation between different uncertainties and visualized an example of the position uncertainties in \Fig\ref{fig:diff_uncer_cor}.
It can be seen that different uncertainties show similar trend across images, but the sampling uncertainties tend to be higher than the uncertainties computed from the first order and Monte Carlo methods (notice different axis scales).
Therefore, using the first order and Monte Carlo uncertainties as error thresholds will under-estimate the localization performance as well (even worse than the sampling uncertainties) and thus should not be used as accuracy metrics.

\PAR{Maximum reprojection difference.} 
Our reference poses are obtained by minimizing a reprojection error in image space, rather than an error in camera pose space. 
Thus, evaluating localization algorithms based on the quality of their reprojections seems a natural metric, especially if these algorithms compute poses by minimizing an image space error. 

For each 3D point in the inlier 2D-3D matches of the reference poses, we compute a reprojection difference between the reference and an estimate pose. 
For each image, we report the maximum difference and we compute the percentages of images that have a maximum reprojection difference below 10, 20, 50 and 100 pixels. 
Since all nighttime images have a resolution of 1600$\times$1200 pixels, these thresholds correspond to 0.5\%, 1\%, 2.5\%, and 5\% of the image diagonal.

Comparing the results with the pose error metric using fixed thresholds, we can see that although the top performing algorithms achieve approximately $90\%$ in the finest pose error category, they only have $70 - 80 \%$ of all the images that were localized within 20 pixel according to the maximum reprojection difference. 
Even less images are localized within 10 pixels. 
Since the accuracy of local features are typically below 5 pixel (cf. \Fig\ref{fig:ablation_ftr_compare}(right)), this indicates that there is still much room for improvement on our extended version of the Aachen Day-Night dataset. %
As such, we believe that the maximum reprojection error metric should be the metric of choice for this dataset.

\section{Conclusion}
\label{sec:conclusion}
In this paper, we have considered the problem of creating reference camera poses for long-term visual localization benchmark datasets. 
In this setting, classical features often struggle to obtain matches between images taken under strongly differing conditions. 
At the same time, human annotations are both time-consuming to generate and not necessarily highly accurate. 
Thus, we have presented an approach for refining reference poses based view synthesis and learned features that allow robust feature matching between real and rendered images. 
In addition, we have discussed multiple metrics for evaluating localization performance. 

The main contribution of this paper is an extensive set of experiments. 
We have shown that the original nighttime reference poses of the Aachen Day-Night dataset are rather inaccurate. 
As a result, the localization accuracy of state-of-the-art methods is currently drastically under-estimate. 
Using our approach, we have created a more accurate set of reference poses. 
We will integrate these poses into the online evaluation service provided at \href{https://www.visuallocalization.net/}{visuallocalization.net} as to provide better evaluations to the community. 
We also used our approach to create an extended version of the Aachen Day-Night dataset and showed that this dataset offers room for improvement. 
We will make the dataset available on the benchmark website. 
Furthermore, we will release the code for our approach as to allow other researchers to more easily build localization benchmarks.

One disadvantage of our approach is its rather slow run-time, taking about 10-20 seconds per iteration for a single image, where most of the time is spend for rendering and especially for the SfM inversion process.
This is not an issue when creating reference poses for a benchmark, as these calculations only need to be done once and can be done offline. 
At the same time, our approach can be used as a post-processing step for any visual localization algorithm. 
An interesting research question is whether more efficient rendering techniques can be used to improve its run-time to a degree that enables online operation. 

\PAR{Acknowledgements.} 
This work was supported by the Swedish Foundation for Strategic Research (Semantic Mapping and Visual Navigation for Smart Robots), the Chalmers AI Research Centre (CHAIR) (VisLocLearn), the National Centre of Competence in Research (NCCR) Robotics, through the Swiss National Science Foundation and the SNSF-ERC Starting Grant, and the European Regional Development Fund under IMPACT No.~CZ.02.1.01/0.0/0.0/15 003/0000468. This article is part of the RICAIP project that has received funding from the European Union's Horizon 2020 research and innovation programme under grant agreement No 857306. 
We thank Mihai Dusmanu and Martin Humenberger for contributing the D2-Net and R2D2 results, respectively.

\bibliographystyle{unsrt}

\bibliography{rpg,references,torsten}   %

\end{document}